\RequirePackage{fix-cm}
\documentclass[twocolumn,envcountsect]{svjour3}          
\smartqed  
\usepackage{amsmath,amsfonts,amssymb,mathtools}
\usepackage{graphics}
\usepackage{epsfig}
\usepackage{threeparttable}
\usepackage{xspace}
\usepackage{siunitx}
\usepackage{times}
\usepackage{graphicx}
\usepackage{subfig}
\usepackage{placeins}
\usepackage{blindtext}
\usepackage{threeparttable}
\usepackage{color}
\usepackage[normalem]{ulem}
\usepackage{multirow}
\usepackage{float}
\usepackage{array}
\usepackage{tabulary}	
\usepackage{microtype}
\usepackage{booktabs}
\usepackage{xspace}
\usepackage[table]{xcolor}
\usepackage{colortbl}
\usepackage{url}
\usepackage{cite}

\usepackage[pagebackref=true,breaklinks=true,colorlinks,citecolor=citecolor,bookmarks=false]{hyperref}
\usepackage{tikz}
\usepackage{forest}
\usetikzlibrary{trees, shapes, shadows, arrows.meta, shapes.multipart, chains, positioning}
\usepackage{marvosym,ifsym}
\usepackage{makecell}

\makeatletter
\newcommand{\thickhline}{%
    \noalign {\ifnum 0=`}\fi \hrule height 1pt
    \futurelet \reserved@a \@xhline
}

\DeclareRobustCommand\onedot{\futurelet\@let@token\@onedot}
\def\@onedot{\ifx\@let@token.\else.\null\fi\xspace}

\definecolor{mygreen2}{RGB}{194, 220, 191}
\definecolor{myred2}{RGB}{248, 206, 204}
\definecolor{myblue2}{RGB}{186, 216, 251}
\definecolor{myyellow2}{RGB}{251, 229, 184}
\definecolor{mypurple2}{RGB}{231, 216, 250}

\definecolor{citecolor}{HTML}{0071bc}
\definecolor{scorered}{HTML}{e4485a}
\definecolor{scoreblue}{HTML}{4a7ee8}
\definecolor{scoregreen}{HTML}{80ba0e}

\definecolor{purple0}{HTML}{e9e9f3}
\definecolor{purple}{HTML}{dcdaed}
\definecolor{purple1}{HTML}{bab6da}

\def\eg{\emph{e.g}\onedot} 
\def\ie{\emph{i.e}\onedot} 
 
\def\etc{\emph{etc}\onedot} 
\def\wrt{w.r.t\onedot} 
\def\iid{i.i.d\onedot} 
\def\etal{\emph{et al}\onedot}
\makeatother
\usepackage[capitalize]{cleveref}
\crefname{section}{Sec.}{Secs.}
\Crefname{section}{Section}{Sections}
\crefname{figure}{Fig.}{Figs.}
\Crefname{figure}{Figure}{Figures}
\crefname{table}{Tab.}{Tabs.}
\Crefname{table}{Table}{Tables}
\newcommand{\numberthis}{\addtocounter{equation}{1}\tag{\theequation}}

\DeclareMathOperator*{\argmin}{arg\,min}

\journalname{IJCV}
\begin{document}

\title{Diffusion Model-Based Video Editing: A Survey}
\author{
    Wenhao Sun~\textsuperscript{1} \and 
    Rong-Cheng Tu~\textsuperscript{1, \Letter} \and 
    Jingyi Liao~\textsuperscript{1} \and 
    Dacheng Tao~\textsuperscript{1, \Letter}
}
\institute{
    Wenhao Sun (wenhao006@ntu.edu.sg) \\
    Rong-Cheng Tu (rongcheng.tu@ntu.edu.sg) \\
    Jingyi Liao (jingyi012@ntu.edu.sg) \\ 
    Dacheng Tao (dacheng.tao@ntu.edu.sg) \\ 
    \\
    \textsuperscript{1}~Nanyang Technological University, Singapore\\
    \textsuperscript{\Letter}~Corresponding authors: Rong-Cheng Tu, Dacheng Tao
}

\date{}

\maketitle

\begin{abstract}
The rapid development of diffusion models (DMs) has significantly advanced image and video applications, making "what you want is what you see" a reality.
Among these, video editing has gained substantial attention and seen a swift rise in research activity, necessitating a comprehensive and systematic review of the existing literature.
This paper reviews diffusion model-based video editing techniques, including theoretical foundations and practical applications.
We begin by overviewing the mathematical formulation and image domain's key methods.
Subsequently, we categorize video editing approaches by the inherent connections of their core technologies, depicting evolutionary trajectory.
This paper also dives into novel applications, including point-based and pose-guided human video editing.
Additionally, we present a comprehensive comparison using our newly introduced V2VBench.
Building on the progress achieved to date, the paper concludes with ongoing challenges and potential directions for future research. 
A project related to this paper can be found at \url{https://github.com/wenhao728/awesome-diffusion-v2v}.
\end{abstract}

\section{Introduction}
Recent years have witnessed remarkable development in Artificial Intelligence Generated Contents (AIGC), catalyzed by high-quality datasets and efficient computational infrastructure.
Particularly in computer vision, AI's impressive capability to generate high-dimensional perceptual data has attracted interest from researchers and exerted a profound significance on industrial products and daily life.
Diffusion models~\cite{jen+15} have emerged as a leading approach for vision generation tasks, including Text-to-Image (T2I) generation~\cite{apa+22,apac+22,cws+22,rad+22,wxx+22,ysx+22,zzx+23} and Image-to-Image (I2I) translation~\cite{gtj22,cwh+22,mfk+23,bso+23,hmt+22,tgb+23,cwq+23,bhe23}.
These methods can generate high-quality and accurate images from text descriptions or modify input images based on specified conditions.

Given that videos are intrinsically composed of continuous image sequences, extending the success of diffusion models from 2D image generation to 3D video generation and editing has sparked considerable research interest.
However, adapting architectural designs for static images to handle the dynamic and temporal aspects of videos is a complex challenge. The scarcity of high-quality video datasets adds further technical difficulties.
Researchers must explore effective video dataset curation, alternative surrogate tasks, and other potential solutions.

\begin{figure*}[htbp]
    \centering
    \includegraphics[width=\linewidth]{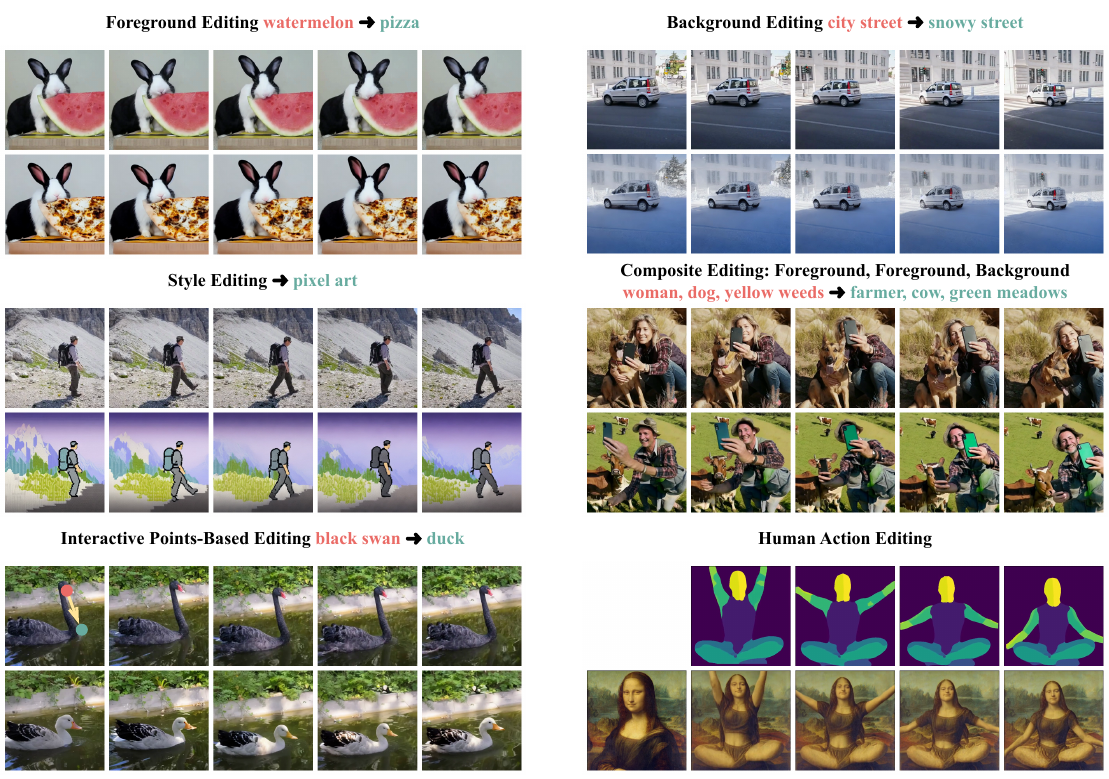}
    \caption{
        This survey focuses on \textbf{video editing} tasks.
        The edited results are obtained using ~\cite{ymc+23,mos+23,CeylanHM23,ryj+23,yyb+23,zjj+23}. We will discuss these methods in \cref{sec:desc-based} and present more experimental results in \cref{sec:bench}.
    }
    \label{fig:task}
\end{figure*}

Despite these challenges, recent advancements have been witnessed in diffusion model-based video generative methods~\cite{jta+22,jwc+22,arh+23,yca+23,xyg+24,rma+23}. Among these methods, the most influential areas are generating videos from text inputs and editing existing videos generatively.
Furthermore, editing existing videos does not require prohibitively expensive video pre-training and allows fine-grained control of the source video, leading to diverse applications, as illustrated in \cref{fig:task}.
At the same time, the increasing fidelity, controllability, and accessibility of diffusion-based video editing raise emerging safety and security concerns.

As the field of diffusion-based video editing continues to expand, there is a growing need for a detailed survey that helps readers stay informed about recent advancements. 
However, existing surveys on diffusion models primarily focus on other aspects or tasks, such as the theoretical formulation of diffusion processes~\cite{hcz+22,zgh23,ang22,lzy+24}, image generation~\cite{ccm+23,tzj+23}, image editing~\cite{yjy+24}, and 3D object generation~\cite{rwv+23,jxt+24}.
While some surveys mentioned video editing~\cite{rwv+23,zqhq+23}, they only provide a brief overview, missing influential methods, detailed narration, and the underlying linkages within these emerging approaches.
This paper provides a thorough review of diffusion model-based video editing technologies. It covers a broad spectrum of editing tasks, methodologies, and evaluations.
Beyond deepening the understanding of the current landscape, this paper also outlines potential developments and applications, highlighting directions for future research.
In summary, the contributions of this paper are as follows:
\begin{itemize}
    \item[$\bullet$]  We present a comprehensive review of diffusion-based video editing studies, encompassing over two hundred papers, and highlight their logical development trajectory.
    \item[$\bullet$] We introduce a new video editing benchmark that encompasses four text-guided video editing tasks. Our evaluation of sixteen widely recognized methods addresses the absence of a uniform benchmark.
    \item[$\bullet$] We summarize the open challenges and outline emerging research trends to inspire future research efforts.
\end{itemize}

The rest of this paper is structured as follows:
In \cref{sec:background}, we review the preliminaries, including the mathematical formulation of diffusion models, related approaches in 2D image generation, and video generation methods.
Subsequently, \cref{sec:desc-based} reviews diffusion model-based video editing approaches and novel applications, categorized into five primary categories.
\cref{sec:bench} introduces V2VBench, a comprehensive video editing benchmark, along with a detailed evaluation and analysis.
In \cref{sec:trend}, we summarize the open challenges and outline emerging research trends.
Finally, \cref{sec:conclusion} summarizes the paper.

\section{Background}\label{sec:background}
This section presents an overview of the essential preliminaries for diffusion model-based video editing methods.
We start with the mathematical framework of diffusion models (\cref{subsec:math}), followed by an exploration of image generation and editing techniques (\cref{subsec:image-diffusion}). 
We then shift to the video domain, providing an overview of diffusion models for video generation and consolidating motion representation in \cref{subsec:video-diffusion}. 
We focus on the discrete formulation of the diffusion process~\cite{jap20} and highlight the core contributions of representative works to maintain focus and conciseness.

\subsection{Mathematicial Framework}\label{subsec:math}
Diffusion models~\cite{jen+15} fall within the family of latent probabilistic models, alongside variational autoencoders (VAEs)~\cite{dm14,ok17}, normalization flows~\cite{ds15},~\etc.
They are distinguished by possessing an analytical form of the latent distribution.

Consider a sample $\mathbf{z}_0$ randomly drawn from the training dataset, with each sample \iid following an underlying distribution $q(\mathbf{z}_0)$. 
The posterior distribution is defined by a Markov chain $q(\mathbf{z}_{1:T}|\mathbf{z}_0) = \prod_{t=1}^{T}q(\mathbf{z}_{t} | \mathbf{z}_{t-1})$, known as the forward diffusion process, which gradually adds Gaussian noise into the data:
\begin{equation}
    \label{eq:ddpm}
    q(\mathbf{z}_{t} | \mathbf{z}_{t-1}) \coloneq 
    \mathcal{N}( s_t \mathbf{z}_{t-1}, \sigma^2_t \mathbf{I})
\end{equation}
where the Markov chain hyperparameters $s_t = \sqrt{{\alpha_t}/{\alpha_{t-1}}}$ and $\sigma_t = \sqrt{1-{\alpha_t}/{\alpha_{t-1}}}$ can be assigned to facilitate the variance preservation formulation~\cite{jap20,ssk+21}. $\{\alpha_t\}_{t=0}^T$ denote a predefined set of hyperparameters that adhere to $0<\alpha_T<\cdots <\alpha_0 = 1$.

The underlying data distribution $q(\mathbf{z}_0)$ is estimated with the form 
$p_{\theta}(\mathbf{z}_0) = \int p_{\theta}(\mathbf{z}_{0:T}) d\mathbf{z}_{1:T}$.
And the joint distribution $p_{\theta}(\mathbf{z}_{0:T})$, known as the reverse diffusion process, is commonly assumed to follow a Markov chain~\cite{jen+15,jap20} with learnable transition distribution:
\begin{equation}
    p_{\theta}(\mathbf{z}_{t-1}|\mathbf{z}_{t}) \coloneq 
    \mathcal{N}\big(
        \mu_{\theta}(\mathbf{z}_{t}, t),
        \Sigma_{\theta}(\mathbf{z}_{t}, t)
    \big)
\end{equation}
where $\theta$ represents the trainable parameters.

\subsubsection{Diffusion Model Properties}
Before diving into diffusion model optimization and inference, let us review some fruitful properties that enhance theoretical derivation and operational efficiency.

\vspace{6pt}
\noindent$\bullet$~\textbf{Gaussian Marginal Distribution.}
The marginal distribution of the forward diffusion process $q(\mathbf{z}_{t} | \mathbf{z}_{0})$ follows Gaussian \cite{jen+15}:
\begin{equation}
    \label{eq:ddpm-t-step}
    q(\mathbf{z}_{t} | \mathbf{z}_{0}) \coloneq 
    \mathcal{N}(\sqrt{{\alpha}_t} \mathbf{z}_{0}, (1-{\alpha}_t) \mathbf{I})
\end{equation}
This implies that the forward diffusion process can be achieved by directly interpolating between Gaussian noise $\epsilon_t \sim \mathcal{N}(\mathbf{0}, \mathbf{I})$ and the initial latent (noise-free sample) $\mathbf{z}_0$ without considering intermediate latents. When $\alpha_T$ is sufficiently closed to 0, $q(\mathbf{z}_{T} | \mathbf{z}_{0})$ converges towards a standard Gaussian distribution $\mathcal{N}(\mathbf{0}, \mathbf{I})$ for arbitray $\mathbf{z}_{0}$.

\vspace{6pt}
\noindent$\bullet$~\textbf{Score Function.}
The score function is defined as the gradient of the log-density~\wrt the data vector~\cite{Hyvarinen05}. Assuming a Gaussian distribution, the score takes the following form~\cite{ssk+21}:
\begin{equation}
    \label{eq:score}
    \nabla_{\mathbf{z}_{t}} \log q(\mathbf{z}_{t} | \mathbf{z}_{0}) 
    = -\frac{\mathbf{z}_{t} - \sqrt{\alpha_t}\mathbf{z}_{0}}{1-\alpha_t}
    = -\frac{\epsilon_t}{\sqrt{1-\alpha_t}}
\end{equation}
where $\epsilon_t \sim \mathcal{N}(\mathbf{0}, \mathbf{I})$ is a standard Gaussian random variable.

\subsubsection{Diffusion Model Optimization}
The diffusion models can be trained by optimizing the Evidence Lower Bound (ELBO)~\cite{jap20}:
\begin{equation}
    \mathrm{ELBO} =
    \mathbb{E}_{q} \Big[
        \log{\frac{p_{\theta}(\mathbf{z}_{0:T})}{q(\mathbf{z}_{1:T} | \mathbf{z}_0)}}
    \Big]
\end{equation}

After parameterization, the model training can be represented as an $\epsilon$-prediction task~\cite{jap20,ys19}:
\begin{equation}
    \label{eq:dm-loss}
    \mathcal{L} = \mathbb{E}_{
        t,\mathbf{z}_t, 
        \epsilon_t\sim\mathcal{N}(\mathbf{0},\mathbf{I})
    }\big[ \lambda(t)\|
        \epsilon_t - \epsilon_\theta (\mathbf{z}_t, t)
    \|_2^2 \big]
\end{equation}
where $\lambda(t) > 0$ serves as a weighting term dependent on the diffusion step $t$. Ho~\etal~\cite{jap20} discovered that setting $\lambda(t) = 1$ in \cref{eq:dm-loss} enhances network learning by emphasizing the more challenging instances with larger $t$. Consequently, they simplified \cref{eq:dm-loss} to:
\begin{equation}
    \label{eq:dm-simple-loss}
    \mathcal{L}_{\mathrm{simple}} = \mathbb{E}_{
        t,\mathbf{z}_t, \epsilon_t
    }\big[ \|
        \epsilon_t - \epsilon_\theta (\mathbf{z}_t, t)
    \|_2^2 \big]
\end{equation}

\subsubsection{Reverse Diffusion Sampling}
The reverse diffusion sampling can be formulated~\cite{jap20} as:
\begin{equation}
    \label{eq:ddpm-samping}
    \mathbf{z}_{t-1} = \sqrt{\alpha_{t-1}}{\mathbf{z}}_{t\rightarrow 0} +
        \sqrt{1 - \alpha_{t-1} - \sigma^2_t} \epsilon_{\theta}(\mathbf{z}_t, t) + \sigma_t \nu_t
\end{equation}
where ${\mathbf{z}}_{t\rightarrow 0} = (\mathbf{z}_t - \sqrt{1 - \alpha_t} \epsilon_{\theta}(\mathbf{z}_t, t))/\sqrt{\alpha_t}$, $\nu_t \sim \mathcal{N}(\mathbf{0}, \mathbf{I})$ denotes standard Gaussian random variable that is independent of $\mathbf{z}_t$, and $\sigma_t = \sqrt{{(1-\alpha_{t-1})}/{(1-\alpha_{t}})} \cdot \sqrt{1-{\alpha_{t}}/{\alpha_{t-1}}}$. 

One practical limitation of \cref{eq:ddpm-samping} is its requirement for $T$ denoising interactions, which is typically set to a large value (\eg $T = 1000$) to conform to the Gaussian assumption.
To accelerate the reverse sampling process, DDIM~\cite{sme20} proposes representing the diffusion process through a non-Markov chain, while retaining the identical objective function as in \cref{eq:dm-loss}.
This non-Markov process allows for the flexibility to sample fewer steps  (\eg $T = 50$) without altering the pre-trained weights. DDIM sampling follows a similar sampling formulation as \cref{eq:ddpm-samping}, but it is a deterministic process with $\sigma_t=0$:
\begin{equation}
    \label{eq:ddim}
    \mathbf{z}_{t-1} = \sqrt{\alpha_{t-1}}{\mathbf{z}}_{t\rightarrow 0} +
        \sqrt{1 - \alpha_{t-1}} \epsilon_{\theta}(\mathbf{z}_t, t)
\end{equation}

\subsection{Image Diffusion Models}\label{subsec:image-diffusion}
This section primarily discusses the image diffusion methods. In practical implementations, the diffusion process often occurs in a semantically equivalent space rather than the original data space (~\eg the VAE latent space versus the pixel space of an image dataset). To avoid ambiguity, we denote the original image samples as $\mathbf{x}_0$ and the noise-free samples in the diffusion process as $\mathbf{z}_0$ throughout the paper.

Early image diffusion models~\cite{ys19,jap20,ap21,pa21,hs22,kaa+22} directly generate images at the desired resolution,~\ie $\mathbf{z}_0 = \mathbf{x}_0$.
As a result, generating high-resolution images is prohibitively expensive, and early image diffusion models typically produce low-resolution images. 

To address the high-resolution challenge, Cascaded Diffusion Models (CDMs)~\cite{apac+22,cws+22,jcw+22} initially generate images at lower resolutions and then upsample them through a super-resolution step. In this scenario, the samples for the base diffusion models training are derived from the downsampled pixel-level images $\mathbf{z}_0 = \mathrm{DownSample}(\mathbf{x}_0)$.

\begin{figure}[tbp]
    \centering
    \includegraphics[trim=0 705 305 2, clip, width=\linewidth]{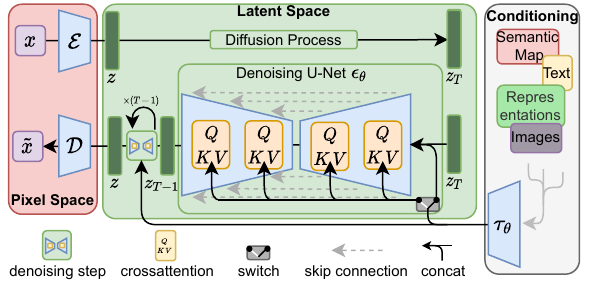}
    \caption{The architecture of LDM~\cite{rad+22}. It consists of an encoder $\mathcal{E}$, a decoder $\mathcal{D}$, and a UNet noise predictor $\epsilon_\theta$ with cross-attention blocks to incorporate conditioning from domain encoders $\tau_\theta$.
    Figure adapted from LDM~\cite{rad+22}.}
    \label{fig:ldm}
\end{figure}

More recently, Latent Diffusion Models (LDMs)~\cite{rad+22,sdj+22} are able to generate high-resolution images at affordable computational cost by operating within the latent space of a pre-trained VAE~\cite{dm14,ok17,prb21}. This approach capitalizes on the observation: VAEs excel at perceptual compression by effectively removing and reconstructing high-frequency details; diffusion models excel at capturing the semantic and conceptual composition of the data~\cite{rad+22}. Specifically, people train diffusion models in the VAEs' lower-dimensional latent space: $\mathbf{z}_0 = \mathcal{E}(\mathbf{x}_0)$, where $\mathcal{E}$ represents a VAE encoder. During inference, the diffusion model outputs are mapped back to the pixel space ${\mathbf{x}}_0 = \mathcal{D}({\mathbf{z}}_0)$ by the VAE decoder $\mathcal{D}$.
The architecture proposed by Rombach~\etal~\cite{rad+22} is illustrated in \cref{fig:ldm}.

\subsubsection{Conditioning and Guidance}\label{subsubsec:condition}
Conditional generation has emerged as one of the most important tasks for image generation. Common conditions include class labels~\cite{pa21,ap21,ssk+21,jap20}, text~\cite{apa+22,jcw+22,apac+22,cws+22,rad+22,sdj+22,wxx+22}, depth maps~\cite{zra23,cxl+24},~\etc.
Then, we survey the architecture design to incorporate conditions and the guidance mechanisms.
To avoid confusion, we explicitly declare the conditional input in the noise predictor: $\epsilon_{\theta}(\mathbf{z}_{t}, t, y)$ represents the noise prediction with condition $y$; $\epsilon_{\theta}(\mathbf{z}_{t}, t, \varnothing)$ represents the unconditional prediction.

\vspace{6pt} \noindent$\bullet$~\textbf{Spatial-Aligned Conditions.}
Many condition variants are spatial aligned with the target image, including depth maps, segmentation maps, and spatial layouts. One straightforward solution is to concatenate these conditions with the noisy latent state $\mathbf{z}_t$ and pass them as input to the noise predictor $\epsilon_\theta$. 
Various I2I translation tasks, such as inpainting and colorization, can be treated as conditional image generation and addressed using this approach~\cite{cwh+22,rad+22}.

Researchers later propose to use the adapter to efficiently handle various spatial alignment conditions without modifying the pre-trained diffusion model parameters.
Two iconic works, T2I-Adapter~\cite{cxl+24} and ControlNet~\cite{zra23}, offer flexible solutions by adding hyper-networks to the pre-trained model. Specifically, they propose duplicating the encoder blocks of the pre-trained UNet noise predictor and reintegrating their outputs.
T2I-Adapter adds the multi-scale hyper-network outputs to the UNet encoder blocks. 
Conversely, the ControlNet outputs are connected to the decoder blocks by zero convolutions to ensure stability in the initial training steps.
Since the parameters remain unchanged, a single pre-trained network can be integrated with multiple adapters for various tasks.

\vspace{6pt} \noindent$\bullet$~\textbf{Adaptive Normalization.} 
Another approach to incorporate conditions into diffusion models is applying the feature-wise transformation~\cite{efh+18}:
$\mathrm{AdaNorm}(\mathbf{h}) = s_y \mathrm{Norm}(\mathbf{h}) + b_y$,
where $s_y \in \mathbb{R}$ and $b_y \in \mathbb{R}$ are derivated from condition $y$.
Group Normalization~\cite{yk18} and Instance Normalization~\cite{dav16,xs17} are common options for hidden feature normalization.
This approach has empirically demonstrated promising results for the time steps and class labels~\cite{ap21,pa21}.

\vspace{6pt} \noindent$\bullet$~\textbf{Conditioning by Cross Attention.}
Many advanced diffusion models \cite{apa+22,cws+22,rad+22} incorporate transformer blocks \cite{ann+17} in their networks and use cross-attention to integrate conditions.
A domain-specific encoder $\tau_{\theta}(\cdot)$ first encodes the condition $y$: $\tau_{\theta}(y) \in \mathbb{R}^{d_{\tau} \times L}$. Then, the cross-attention layer $\mathrm{Attn}(\mathbf{Q}, \mathbf{K}, \mathbf{V})$ is formulated as:
\begin{align}
    \label{eq:corss-qkv}
    &\mathbf{Q} = \mathbf{W}_Q \cdot \mathbf{h} ,\ 
    \mathbf{K} = \mathbf{W}_K \cdot \tau_{\theta}(y),\ 
    \mathbf{V} = \mathbf{W}_V \cdot \tau_{\theta}(y)\\
    &\mathrm{Attn}(\mathbf{Q}, \mathbf{K}, \mathbf{V}) = \mathrm{Softmax}(\mathbf{Q}^T \cdot \mathbf{K}/\sqrt{d}) \cdot \mathbf{V}^T
\end{align}
where $\mathbf{h} \in \mathbb{R}^{d \times N}$ denotes the flattened hidden feature of the noise predictor $\epsilon_{\theta}(\cdot)$. $\mathbf{W}_Q \in \mathbb{R}^{d \times d}$, $\mathbf{W}_K \in \mathbb{R}^{d \times d_{\tau}}$, and $\mathbf{W}_V \in \mathbb{R}^{d \times d_{\tau}}$ denote learnable projection matrices.
Text descriptions and reference images are common cross-attention conditions.

\vspace{6pt} \noindent$\bullet$~\textbf{Classifier Guidance.}
In addition to the model inputs, Sohl-Dickstein~\etal~\cite{jen+15} and Song~\etal~\cite{ssk+21} propose using pre-trained classifier $\mathcal{C}(\cdot)$, which predicts the class probability of the intermediate diffusion latents, to guide the generation process of diffusion models. 
They formulate the estimated score function as follows:
\begin{align*}
    \nabla_{\mathbf{z}_{t}} &\log p_{\theta}(\mathbf{z}_{t}, y)
    = \nabla_{\mathbf{z}_{t}} \log p_{\theta}(\mathbf{z}_{t}) +      
        \nabla_{\mathbf{z}_{t}} \log \mathcal{C}(y | \mathbf{z}_{t}) \numberthis \\
    &= -\frac{1}{\sqrt{1-\alpha_t}} \big( 
            \epsilon_{\theta}(\mathbf{z}_{t}, t, \varnothing) - 
            \sqrt{1-\alpha_t} \nabla_{\mathbf{z}_{t}} \log \mathcal{C}(y | \mathbf{z}_{t})
        \big)
\end{align*}
Thus, a new classifier-guided noise predictor takes the following form:
\begin{equation}\label{eq:cbg}
    {\epsilon}_{\theta}(\mathbf{z}_{t}, t, y) \leftarrow
        \epsilon_{\theta}(\mathbf{z}_{t}, t, \varnothing) - 
        w \sqrt{1-\alpha_t} \nabla_{\mathbf{z}_{t}} \log \mathcal{C}(y | \mathbf{z}_{t})
\end{equation}
where the scale $w$ is added to control the guidance strength.

\vspace{6pt} \noindent$\bullet$~\textbf{Classifier-Free Guidance.}
Ho \& Salimans~\cite{hs22} propose guiding the diffusion generation process without pre-trained classifiers.
They apply Bayes' theorem to formulate the classifier gradient $\nabla_{\mathbf{z}_{t}} \log \mathcal{C}(y | \mathbf{z}_{t})$ using the conditional score function and unconditional score function:
\begin{align*}
    \nabla_{\mathbf{z}_{t}} & \log \mathcal{C}(y | \mathbf{z}_{t})
    = \nabla_{\mathbf{z}_{t}} \log p_{\theta}(\mathbf{z}_{t} | y) -  
        \nabla_{\mathbf{z}_{t}} \log p_{\theta}(\mathbf{z}_{t})
         \numberthis\\
    &= -\frac{1}{\sqrt{1-\alpha_t}} \big( 
            \epsilon_{\theta}(\mathbf{z}_{t}, t, y) - \epsilon_{\theta}(\mathbf{z}_{t}, t, \varnothing)
        \big)
\end{align*}
which is substituted into \cref{eq:cbg}, resulting in a new classifier-free predictor:
\begin{align*}
    {\epsilon}_{\theta}(\mathbf{z}_{t},& t, y) 
    \leftarrow \epsilon_{\theta}(\mathbf{z}_{t}, t, \varnothing) 
        - w \sqrt{1-\alpha_t} \nabla_{\mathbf{z}_{t}} \log \mathcal{C}(y | \mathbf{z}_{t})\\
    &= \epsilon_{\theta}(\mathbf{z}_{t}, t, \varnothing) + w \big( 
            \epsilon_{\theta}(\mathbf{z}_{t}, t, y) - \epsilon_{\theta}(\mathbf{z}_{t}, t, \varnothing)
        \big)\numberthis\label{eq:classifier-free}
\end{align*}

During training, the condition $y$ is randomly set to empty $\varnothing$, allowing a single predictor to handle both conditional and unconditional inputs.

This classifier-free guidance can naturally extend to multiple conditions with a hierarchical structure. Here, we illustrate the scenario with two conditions; additional scenarios can be similarly extended:
\begin{align*}
    {\epsilon}_{\theta}(\mathbf{z}_{t},& t, y_1, y_2)  
    \leftarrow \epsilon_{\theta}(\mathbf{z}_{t}, t, \varnothing, \varnothing)\\
    & + w_1 \big( 
        \epsilon_{\theta}(\mathbf{z}_{t}, t, y_1, \varnothing) 
        - \epsilon_{\theta}(\mathbf{z}_{t}, t, \varnothing, \varnothing)
    \big)\numberthis\label{eq:multi-cfg}\\
    & + w_2 \big( 
        \epsilon_{\theta}(\mathbf{z}_{t}, t, y_1, y_2) 
        - \epsilon_{\theta}(\mathbf{z}_{t}, t, y_1, \varnothing)
    \big)
\end{align*}

\subsubsection{Editing and Customization}\label{subsubsec:img-editing}
Closely related to conditional generation, image editing concentrates on generating images while maintaining specific attributes, such as the identity of subjects, background, spatial layout,~\etc.
Here, we discuss image editing techniques beyond the previously mentioned conditioning mechanisms.

\vspace{6pt} \noindent$\bullet$~\textbf{Diffusion Latent Initialization.}
Image diffusion models typically initialize the latent states with Gaussian noise~\cite{jen+15,ys19,jap20,kaa+22}. To preserve low-frequency structures while editing high-frequency details, SDEdit~\cite{cyy+22} perturbs the source image through the forward diffusion process, as indicated in \cref{eq:ddpm-t-step}, then generates the target image from the partially perturbed latent state. Many feasible images can correspond to the same initial latent state, and conditions and guidance techniques from \cref{subsubsec:condition} are employed to direct the editing process~\cite{gtj22}.

\vspace{6pt} \noindent$\bullet$~\textbf{Diffusion Latent Inversion.}
Instead of perturbing the source image with randomness, the latent state can be initialized using the reverse diffusion process~\cite{sme20}.
Specifically, the sampling formula in \cref{eq:ddim} is algebraically manipulated as follows:
\begin{equation}
    \label{eq:ddim_rev}
    \mathbf{z}_t = 
        \sqrt{\frac{\alpha_t}{\alpha_{t-1}}}\mathbf{z}_{t-1} +
        \bigg(
            \sqrt{1-\alpha_{t}} - \sqrt{\frac{\alpha_t}{\alpha_{t-1}}-\alpha_{t}}
        \bigg)\epsilon_\theta(\mathbf{z}_{t}, t, y)
\end{equation}
Here, $\mathbf{z}_{t}$ appears on both sides of \cref{eq:ddim_rev}.
Assuming Gaussian transitions and small step sizes, $\mathbf{z}_t$ can be approximated by replacing $\epsilon_\theta(\mathbf{z}_{t}, t)$ on the right-hand side of \cref{eq:ddim_rev} with $\epsilon_\theta(\mathbf{z}_{t-1}, t)$, resulting in the following approximation:
\begin{equation}
    \label{eq:ddim_inv}
    \mathbf{z}_t \approx
        \sqrt{\frac{\alpha_t}{\alpha_{t-1}}}\mathbf{z}_{t-1} +
        \bigg(
            \sqrt{1-\alpha_{t}} - \sqrt{\frac{\alpha_t}{\alpha_{t-1}}-\alpha_{t}}
        \bigg)\epsilon_\theta(\mathbf{z}_{t-1}, t, y)
\end{equation}
Then, reverse diffusion sampling on the inverted latent states leads to deterministic reconstruction with negligible error when using the inversion condition $y$, and enables editing when using the target condition $\hat{y}$.

When incorporating classifier-free guidance~\cite{hs22}, the $w>1$ guidance scale amplifies the approximation error, which further accumulates throughout the diffusion process. 
Null-text inversion~\cite{mha+23} addresses this issue by regressing the `null text' embedding at each denoising step. For each timestep $t=T,\ldots, 1$, the null-text embeddings can be iteratively optimized using the following formula:
\begin{equation}
    \label{eq:null_inv}
    \hat{\varnothing}_t = \argmin_{\varnothing_t} \| 
    \mathbf{z}_{t-1}(\mathbf{z}_{t}, t, y, \varnothing_t) - \mathbf{z}^u_{t-1}
    \|_2^2
\end{equation}
where the latent variable $\mathbf{z}_{t-1}(\mathbf{z}_{t}, t, y, \varnothing_t)$ is obtained from the inversion process \cref{eq:ddim_inv} with classifier-free noise prediction as described in \cref{eq:classifier-free}.
And $\mathbf{z}^u_{t-1}$ is obtained from the unconditional inversion ($y=\varnothing$).

\begin{figure}[t]
    \centering
    \includegraphics[width=\linewidth]{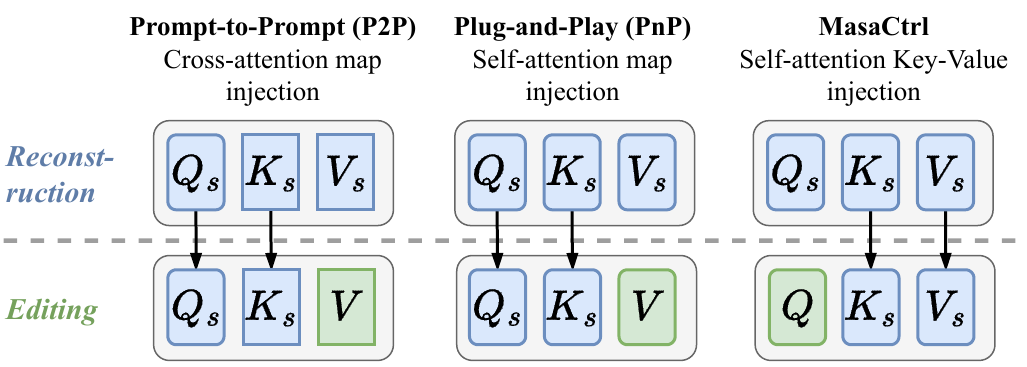}
    \caption{
    Three representative feature injection methods for image editing. P2P~\cite{hmt+22} injects cross-attention maps from the reconstruction branch into the editing branch. PnP~\cite{tgb+23} injects self-attention maps. While MasaCtrl~\cite{cwq+23} injects self-attention query and key features.}
    \label{fig:feat-inject}
\end{figure}

\vspace{6pt} \noindent$\bullet$~\textbf{Attention Feature Injection.}
The network attention maps and hidden features possess abundant semantic and spatial information~\cite{tjw+23}.
Several methods focus on denoising with dual branches in parallel and selectively inject attention maps and hidden features from one branch to another, ensuring consistency and editability. Specifically, this paradigm employs a reconstruction branch for source images and an editing branch for target images as illustrated in \cref{fig:feat-inject}.

Prompt-to-Prompt (P2P)~\cite{hmt+22} emphasizes the efficacy of text prompts in T2I diffusion models. It injects the cross-attention map of unchanged tokens from the reconstruction branch into the editing branch, thereby preserving the unaltered area of source images.
Plug-and-Play (PnP)~\cite{tgb+23} and MasaCtrl~\cite{cwq+23} emphasize the importance of self-attention layers as structural descriptors. PnP injects self-attention query, key, and each block's output from the reconstruction branch to the editing branch, ensuring the preservation of semantic layout.
On the other hand, MasaCtrl maintains the query features unchanged while transferring key and value features. It uses the injection masks by thresholding the cross-attention maps of edited words~\cite{afl23,alf22} to address foreground-background confusion.
Feature injection methods can start by inverting the source image through latent inversion, as shown in \cref{eq:ddim_inv,eq:null_inv}. Subsequently, the inverted latent states are used for both branches.

\vspace{6pt} \noindent$\bullet$~\textbf{Text Inversion.}
Text conditions are pivotal in denoising when noisy images offer limited information during earlier reverse diffusion steps~\cite{ysx+22}.
Textural Inversion~\cite{ryy+23} optimizes a token embedding, serving as an identifier to maintain the protagonist's identity.
DreamBooth~\cite{nyv+23} employs an identifier alongside the subject's class name, capturing semantic information and finer details by fine-tuning the base and super-resolution components of Imagen~\cite{cws+22} using 3-5 images.
Continuing in this direction, subsequent studies \cite{nbr+23,why+23,jwz+23,hyx+23,jyn+23} have achieved precise attribute control and enabled additional applications, such as conceptual interpolation.

Among these image editing methods, latent state initialization methods are proficient in large features and pose changes~\cite{yba+23}. 
Feature injection methods excel in stylization, background, and object replacements, while text inversion methods are more suitable for object customization and personalization. 
Many orthogonal methods can be collaboratively employed to address diverse editing tasks.

\subsubsection{Efficient Adaptions}\label{subsubsec:peft}
\noindent$\bullet$~\textbf{Low-Rank Adaptation (LoRA)}~\cite{eyp+22} is a widely recognized matrix reparameterization technique. For a pre-trained parameter matrix $\mathbf{W} \in \mathbb{R}^{d \times k}$, LoRA introduces two low-rank factorized matrices $\mathbf{A} \in \mathbb{R}^{d \times r}$, $\mathbf{B} \in \mathbb{R}^{r \times k}$, and $r \ll \min{(d,k)}$ to represent its training updates $\Delta \mathbf{W} = \mathbf{A} \cdot \mathbf{B} \in \mathbb{R}^{d \times k}$. Only the factorized matrices, $\mathbf{A}$ and $\mathbf{B}$, are trainable: $\mathbf{W} \leftarrow \mathbf{W} + \mathbf{A} \cdot \mathbf{B}$. 
LoRA can be effectively applied to various network layers, including attention and linear layers, offering advantages such as lower memory consumption and faster convergence.

\vspace{6pt} \noindent$\bullet$~\textbf{Token Merging (ToMe)}~\cite{dcx+23,dj23} is an efficient adaptation to increase the throughput of the Vision Transformer (ViT) architecture \cite{ala+21}. ToMe reduces the number of tokens processed by merging similar tokens before attention layers. This merging operation uses bipartite soft matching, where tokens are alternatively partitioned into two sets, and the top $r$ most similar tokens between these sets are merged.
ToMe later extends to dense prediction tasks, such as image generation. For StableDiffusion~\cite{rad+22}, ToMe happens before self-attention layers, with outputs unmerged to maintain the token count. This approach doubles the image generation speed and reduces memory consumption by up to $5.6 \times$ with negligible quality degradation.

\subsection{Video Diffusion Models}\label{subsec:video-diffusion}
Video generation aims to create videos based on various non-video conditions, sharing many underlying technologies with video editing.
Recent advancements in diffusion models \cite{jhd+23,djj+23,hmy+23,wcm+23,yca+23,smj+23,xyg+24} have accelerated video generation progress, surpassing previous GANs~\cite{clq+21,ism22,vac+22}. 
We next review key technologies in video generation.

\begin{figure}[t]
    \centering
    \includegraphics[width=\linewidth]{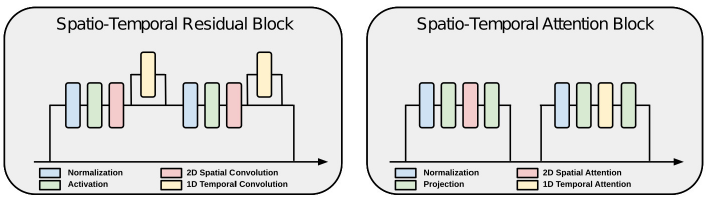}
    \caption{The temporal extension from image to video models involves incorporating 1D temporal convolutions after each 2D spatial convolution (left). Additionally, each 2D spatial attention block is followed by a 1D temporal attention block (right). Figure adapted from Gen-1~\cite{pjp+23}.}
    \label{fig:temporal-layers}
\end{figure}

\begin{figure}[t]
    \centering
    \includegraphics[width=\linewidth]{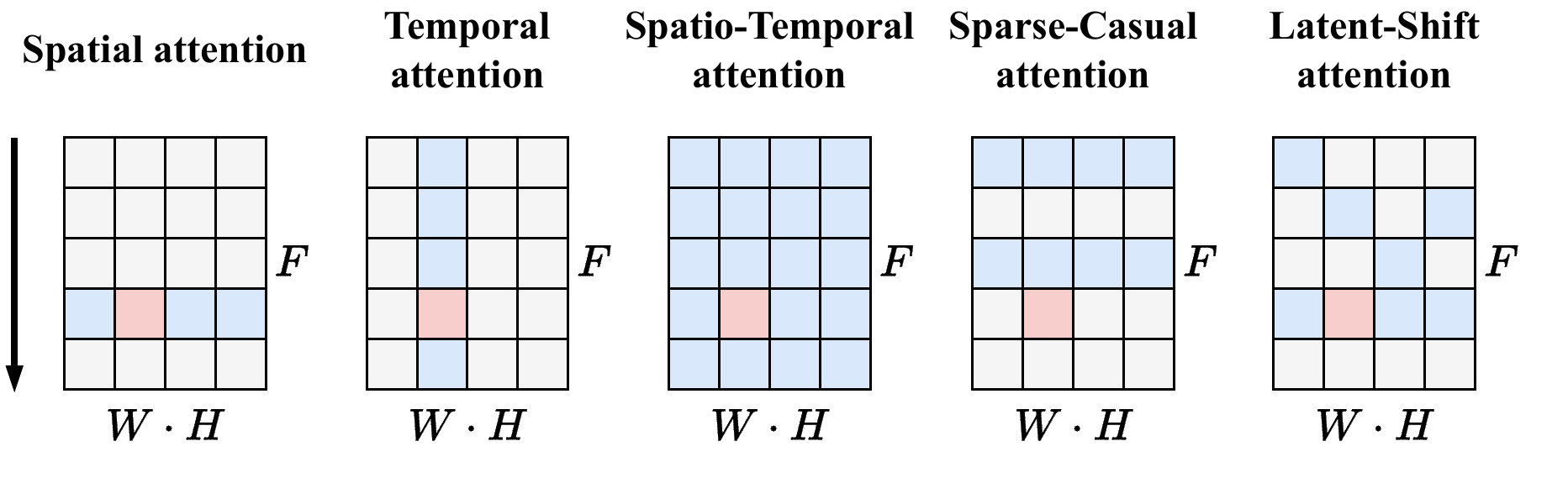}
    \caption{Self-attention variations. The query tokens are in red, and the key and value tokens are in blue. $W$, $H$, and $F$ denote the input video's width, height, and frame numbers.}
    \label{fig:self-attention}
\end{figure}

Pre-trained image diffusion models are well-known for their 2D modeling capabilities. Since video frames are sequential images, extending these image priors to video tasks is a logical progression. The seminal work, Video Diffusion Models (VDM)~\cite{jta+22}, adopts a factorized 3D architecture that separately processes the 2D spatial and 1D temporal dimensions. 
\cref{fig:temporal-layers} illustrates two variants of 3D blocks: temporal convolutions are incorporated through residual connections after each spatial convolution; attention blocks apply a similar architecture.
This design facilitates cross-frame interaction at a relatively low cost, mitigating the computational burden imposed by the curse of dimensions. \cref{fig:self-attention}~also illustrates the distinction between spatial, temporal, and full 3D spatial-temporal attention.

Subsequent studies extend the success of CDM and LDM in the Text-to-Video (T2V) domain on larger data and model scales. Models such as ModelScope~\cite{jhd+23}, Show-1~\cite{djj+23}, VideoCrafter~\cite{hmy+23,hyx+24}, and LaVie~\cite{wcm+23} serve as large-scale open-source baselines, leveraging pre-trained T2I diffusions for efficient training. GenTron~\cite{smj+23} and LATTE~\cite{xyg+24} utilize a pure transformer backbone in video diffusion models. 
Additionally, Latent-Shift~\cite{jsh+23} explores the latent-shift attention (as shown in \cref{fig:self-attention}) in video generation tasks as an alternative implementation for spatiotemporal modeling.
AnimateDiff~\cite{yca+23} showcases the orthogonal learning of appearance and motion. It freezes the original spatial weights while learning adaptable temporal layers on video datasets. 
Once the temporal layers are trained, they can serve as plugins, enabling various customized image diffusion models from the community to gain animation capabilities.

In addition to T2V, Image-to-Video (I2V)~\cite{cwz+23,zwz+23,rma+23,zrn+23} and video completion~\cite{xcj+23,wsv+22} tasks also attract significant research interest. These tasks involve providing a single frame or short video clip to generate longer videos.
Other conditions include audio~\cite{vtn+23,scd+23,yws+23}, structure~\cite{jmy+23,tch+23}, and even MRI signals~\cite{ChenQZ23}. Multi-modal guided video generation~\cite{Zhu0H0TCGSF23,TangYZ0B23,shl+23} is also actively pursued.

\subsection{Video Motion Representation}\label{subsec:optical-flow}
Before discussing video editing methods, let us discuss an iconic video motion representation: dense optical flow \cite{bb81,SandT06}, depicting each pixel's horizontal and vertical offset.
It finds extensive utility in point tracking \cite{SandT06,HarleyFF22}, video compression \cite{VadakitalDLTLR22}, evaluation metrics \cite{zyj+23,yxx+23}, and video editing \cite{yba+23,lww+23,zd23,ymc+23,esj+23,YangZLL23,zj20}.
Many optical flow estimation models~\cite{zj20,XuZ0RT22}, trained and validated using simulated ground-truth datasets~\cite{DosovitskiyFIHH15,MayerIHFCDB16,ButlerWSB12,GaidonWCV16}, exhibit robust generalization to realistic scenarios.

Throughout this paper, we denote the optical flow from the source frame ${\mathbf{x}}^{f_1} \in \mathbb{R}^{3 \times W \times H}$ to the target frame ${\mathbf{x}}^{f_2} \in \mathbb{R}^{3 \times W \times H}$ as $\mathcal{F}^{f_1 \rightarrow f_2} \in \mathbb{R}^{2 \times W \times H}$. 
Certain regions may lack accurate correspondence, often indicated by occlusion masks $\mathcal{O}^{f_1 \rightarrow f_2} \in \{0, 1\}^{W \times H}$. 
The operation of estimating the target frame ${\mathbf{x}}^{f_2}$ from an optical flow $\mathcal{F}^{f_1 \rightarrow f_2}$ and a source frame ${\mathbf{x}}^{f_1}$ is known as warping $\mathcal{F}^{f_1 \rightarrow f_2}(\mathbf{x}^{f_1})$.

\section{Diffusion Model-Based Video Editing}\label{sec:desc-based}
\begin{figure}[t]
    \centering
    \includegraphics[width=\linewidth]{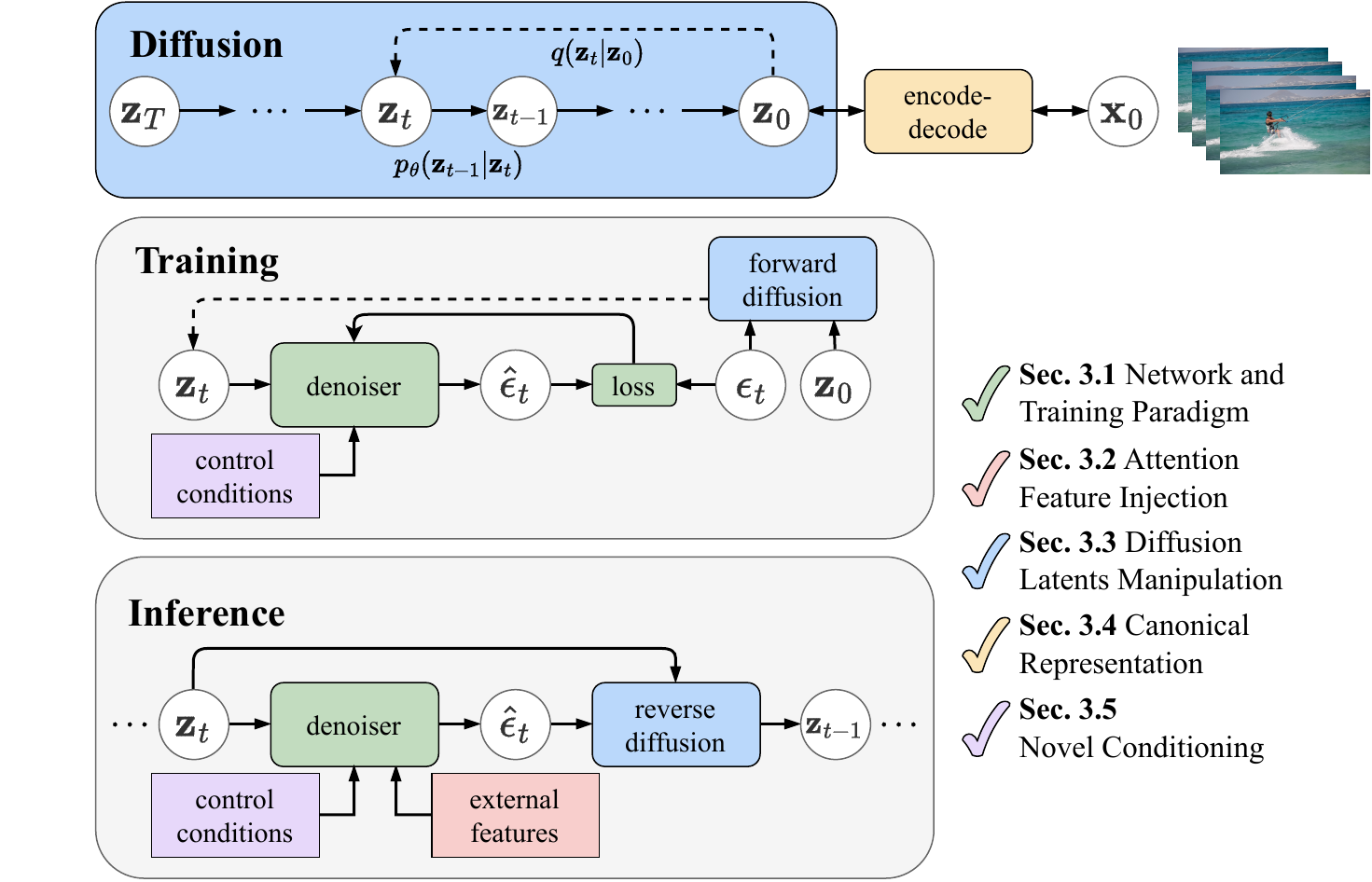}
    \caption{
        Overview of diffusion-based video editing components.
    }
    \label{fig:taxonomy}
\end{figure}

\begin{figure*}
    \centering
    \includegraphics[width=\textwidth]{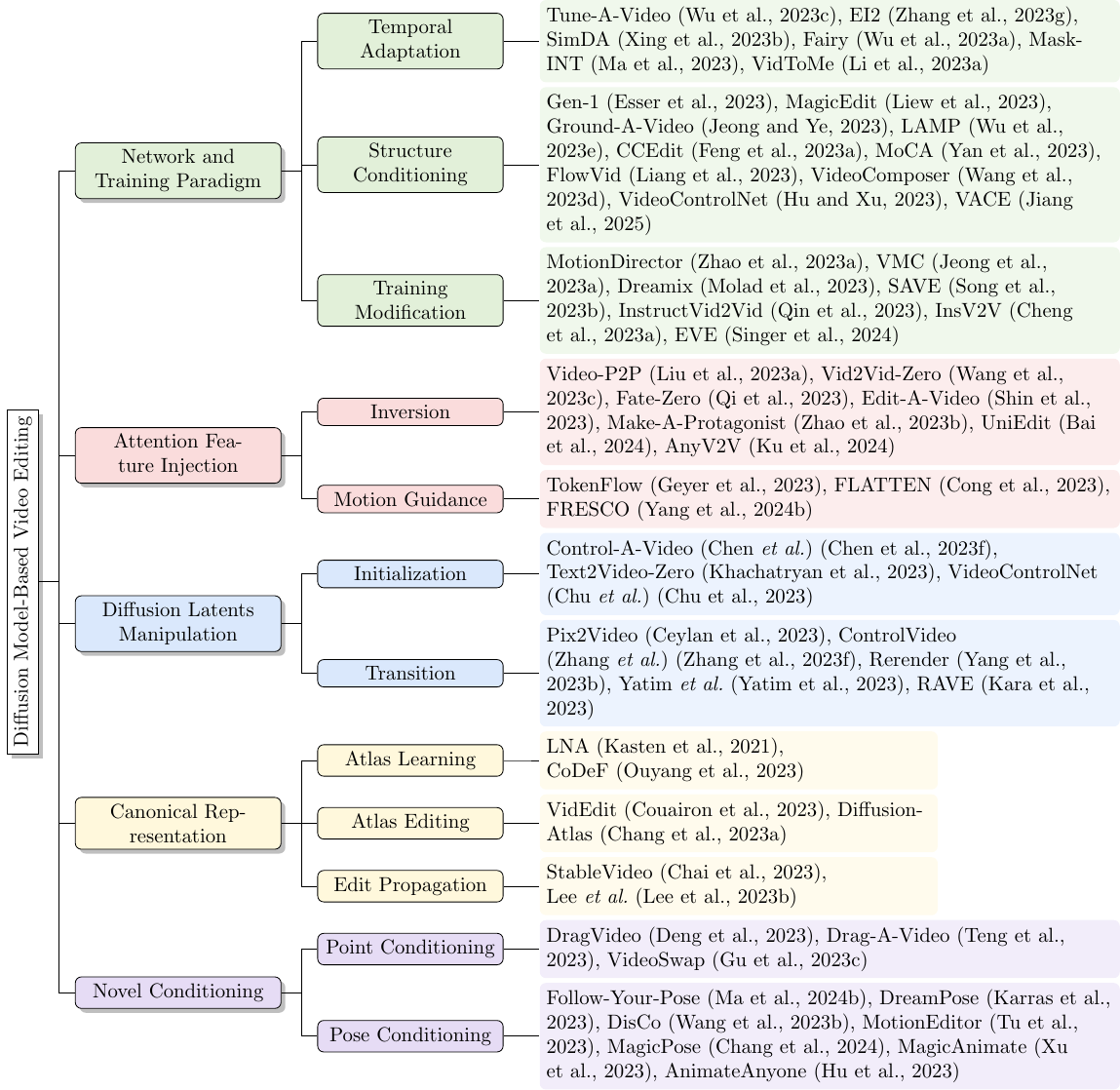}
    \caption{Taxonomy of diffusion-based video editing methods.}
    \label{fig:taxonomy-desc}
\end{figure*}

Building on the generation concept, editing tasks focus on altering existing video content rather than generating new material from scratch.
Video editing tasks, also known as Video-to-Video (V2V) translation, use the source video $\mathbf{x}^s_0$ and the target prompt $y$ serve as inputs to generate the desired changes in the resulting video $\mathbf{{x}}_0$.

\cref{fig:taxonomy} depicts the core components of a general video editing diffusion framework.
We first provide a detailed review of network and training paradigm modifications (\cref{subsec:ino-modules}), spanning initial test-time tuning strategies designed to overfit specific video motion patterns and generalized methods trained on massive datasets for zero-shot video editing.
Moving forward, we revisit methods utilizing attention feature injection (\cref{subsec:feat-inject}) to lift image generation models to the video domain. Rather than requiring the model to memorize dynamics through optimization, these approaches enable motion by injecting features guided by source video motion cues (\eg, optical flow) with minimal or no tuning overhead.
Furthermore, \cref{subsec:diffusion-steps} covers techniques that manipulate diffusion latents, rather than model weights or internal features, to replicate source dynamics.
\cref{subsec:canonical} explores canonical video representation methods, which learn mappings between 3D video volumes and 2D atlases. By establishing mappings between video and image modalities, these methods allow image editing tools to be reused in video scenarios without direct network adaptation.
Finally, we introduce approaches incorporating proactive control conditions (\cref{subsec:control}), such as mouse trajectories and human poses, facilitating more intuitive user interactions.

Certain methods integrate diverse techniques that span multiple categories.
We categorize these based on their primary novel methodology while providing cross-references to other relevant sections.

\subsection{Network and Training Paradigm}\label{subsec:ino-modules}
Applying image editing methods to videos on a frame-by-frame basis introduces temporal inconsistency and quality degradation issues.
This section explores various network and training paradigm modifications, including network layer design, structure conditioning, and training modification. These methods collectively reveal the evolving effectiveness and complexity of video editing tasks.

\subsubsection{Temporal Adaptation}
Tune-A-Video~\cite{wgw+23} pioneers the concept of one-shot tuning in video editing. It transforms the spatial self-attention layers of a T2I StableDiffusion into the sparse-casual attention layers.
As illustrated in \cref{fig:self-attention}, this mechanism employs the key-value features drawn from both the first frame $\mathbf{h}^{1}$ and the preceding frame $\mathbf{h}^{f-1}$ to update the current frame features. The key and value features for frame $f$ are defined as follows:
\begin{equation}
    \mathbf{K}^f = \mathbf{W}_K \cdot [\mathbf{h}^{1}, \mathbf{h}^{f-1}], \mathbf{V}^f = \mathbf{W}_V \cdot [\mathbf{h}^{1}, \mathbf{h}^{f-1}]
\end{equation}
where $[\cdot]$ denotes concatenation along the sequence dimension. $\mathbf{W}_K \in \mathbb{R}^{d \times d}$ and $\mathbf{W}_V \in \mathbb{R}^{d \times d}$ represent the projection matrices.
To effectively capture the structure and motion of the source video, Tune-A-Video fine-tunes the query projection matrices of sparse-causal attention, query projection matrices of cross-attention layers, and the newly introduced temporal attention layers.

EI2~\cite{ZhangLNHGL23} analyzes the covariate shifting caused by the temporal attention layers in Tune-A-Video~\cite{wgw+23}. It identifies a deficiency of the Layer Normalization to adequately center the hidden features $\mathbf{h} \in \mathbb{R}^{d \times F \times W \times H }$ along the temporal dimension.
As a solution, it proposes an Instance Centering (IC) layer as the replacement:
\begin{equation}
    \mathrm{IC}(\mathbf{h}) = \mathbf{h} - \frac{1}{F\cdot W \cdot 
 H}\sum_{f=1}^{F}\sum_{i=1}^{W}\sum_{j=1}^{H} \mathbf{h}^f_{[i,j]}
\end{equation}
Moreover, Spectral Normalization~\cite{mkk+18} is incorporated to mitigate covariate shifts. Furthermore, EI2 enhances its approach by concatenating downsampled inter-frame features as self-attention key and value features, facilitating efficient spatio-temporal interaction.

While EI2~\cite{ZhangLNHGL23} addresses the normalization issues inherent in Tune-A-Video~\cite{wgw+23}, both remain computationally expensive due to the need for full or partial UNet updates. This limitation catalyzed the development of parameter-efficient fine-tuning approaches, which freeze the backbone to preserve the generative prior while inserting lightweight temporal modules.

\vspace{6pt} \noindent$\bullet$~\textbf{Efficient Approaches.}
SimDA~\cite{zqh+23} aims to extend T2I generation to T2V generation by parameter-efficient fine-tuning. It uses an adapter comprising two learnable fully connected (FC) layers: $\mathbf{W}_{\mathrm{down}} \in \mathbb{R}^{d' \times d}$ and $\mathbf{W}_{\mathrm{up}} \in \mathbb{R}^{d \times d'}$, along with an intermediate layer $\mathcal{L}(\cdot)$:
\begin{equation}
    \mathrm{Adapter}(\mathbf{h}) = \mathbf{h} 
        + W_{\mathrm{up}} \cdot \mathcal{L}(W_{\mathrm{down}}\cdot  \mathbf{h})
\end{equation}
The first FC layer maps the input $\mathbf{h} \in \mathbb{R}^{d \times F \times W \times H}$ to a lower-dimension vector space ($d' < d$), while the second layer maps it back. SimDA incorporates spatial adapters to learn appearance transferability and temporal adapters to model temporal information. The intermediate layers utilize GELU activations~\cite{hg16} and depth-wise 3D convolutions for spatial and temporal adapters, respectively.
Additionally, SimDA implements the temporally latent-shifted self-attention~\cite{lgh19,jsh+23}, shown in \cref{fig:self-attention}, to enhance temporal consistency. 
This approach thus benefits from both computational efficiency and temporal perception.

An inefficiency in the aforementioned methods lies in their frame-wise interdependencies. Fairy \cite{bcx+23} addresses this bottleneck by decoupling these dependencies and parallelized frame sampling.
First, some keyframes are sampled in an interleaved manner. These keyframes are edited in a single forward pass, utilizing a 3D spatio-temporal self-attention extended from a T2I diffusion model.
Fairy independently processes the remaining frames using an instruction-based I2I diffusion model~\cite{bhe23}, facilitating parallel processing. The self-attention layers of the I2I model are adjusted to integrate additional key and value features from the keyframes, providing supplementary guidance to ensure temporal consistency.
The I2I model is fine-tuned using affine augmentation, applied identically and randomly to both source and target images, ensuring robustness in movements.

Like Tune-A-Video~\cite{wgw+23}, MaskINT \cite{hsb+23} also edits keyframes using an extended T2I diffusion model with spatio-temporal self-attention \cite{wgw+23}.
While it emphasizes efficiency by deriving intermediate frames through interpolation using non-diffusion models.
Specifically, it employs a masked generative transformer with structural guidance, achieving performance comparable to diffusion-based methods while reducing inference time by requiring fewer iterations.

Instead of building spatio-temporal dependencies through global attention, VidToMe~\cite{xcx+23} leverages temporal redundancy in videos through Token Merging (ToMe)~\cite{dcx+23,dj23}. 
It partitions video frames into chunks and applies inter-frame ToMe within each chunk to reduce the number of self-attention layers input tokens. To ensure long-term consistency, a set of cross-chunk global tokens is maintained, merging with and being updated by each chunk. VidToMe is compatible with various image editing approaches, such as ControlNet~\cite{zra23} and PnP~\cite{tgb+23}.

\subsubsection{Structure Conditioning}
One challenge in video editing is the occurrence of temporal instability, often resulting in distortions in the shapes of objects. Many recent works explicitly use structural representations to address this issue.
Such structural conditioning represents an explicit approach to the disentanglement problem in video editing. By providing a geometric scaffold (\eg, depth or poses), these methods decouple the preservation of the source video's layout from the synthesis of new textures. The field has progressed from dense spatial priors (depth maps), which can be overly restrictive, to sparse/semantic priors (bounding boxes and appearance anchors) that allow for greater creative flexibility.

\vspace{6pt} \noindent$\bullet$~\textbf{Depth Maps.}
Gen-1~\cite{pjp+23} concatenates depth maps~\cite{RanftlLHSK22} with the noisy latents $\mathbf{z}_t$ as inputs to the noise predictor, effectively offering sequential structure guidance. 
Nevertheless, a major limitation of Gen-1 is the high computational cost involved in fine-tuning all parameters for video data.
To address this limitation, MagicEdit~\cite{jhj+23} freezes the pre-trained parameters of the T2I model and fine-tunes the new temporal layers on video data, similar to AnimateDiff~\cite{yca+23}. Subsequently, it directly integrates the ControlNet~\cite{zra23} trained for conditional image generation.
The integrated model demonstrates capabilities in video stylization, local editing, conceptual mixing,~\etc.

\vspace{6pt} \noindent$\bullet$~\textbf{Bounding Boxes.}
Ground-A-Video~\cite{hj23} utilizes depth maps to offer spatial control through an inflated ControlNet~\cite{zra23}. Moreover, Ground-A-Video integrates bounding boxes to explicitly define the regions that should remain unedited. The bounding boxes from GLIP~\cite{LiZZYLZWYZHCG22} are encoded as supplementary key-value features within cross-attention and temporal self-attention layers.

\vspace{6pt} \noindent$\bullet$~\textbf{Appearance Conditions.}
During training, the model inadvertently memorizes appearance information from the source video, resulting in unexpected visual elements in the edited videos.
To address this challenge, LAMP~\cite{rlt+23} proposes editing the first frame using off-the-shelf image editing methods~\cite{sme20,zra23}. Then, the edited frame, along with other frames, is input into the denoising model, effectively decoupling appearance and motion from the learning process. 
LAMP~\cite{rlt+23} also proposes the use of spatio-temporal convolution layers to enhance temporal consistency:
\begin{equation}
    \mathrm{ST}\text{-}\mathrm{Conv}(\mathbf{h}) = 
    \mathrm{Sigmoid}\big(\mathrm{S}\text{-}\mathrm{Conv}(\mathbf{h})\big) \odot \mathrm{T}\text{-}\mathrm{Conv}(\mathbf{h})
\end{equation}
where $\mathbf{h} \in \mathbb{R}^{d \times F \times W \times H}$ represents the hidden feature, with $\mathrm{S}\text{-}\mathrm{Conv}(\cdot)$ and $\mathrm{T}\text{-}\mathrm{Conv}(\cdot)$ operating on its $d \times W \times H$ and $d \times F$ dimensions, respectively. And $\odot$ denotes the element-wise multiplication.
During inference, LAMP adopts a shared latent state initialization strategy~\cite{ssg+23} to enhance consistency.
Specifically, it samples a shared anchor latent state $\mathbf{z}_T^{\mathrm{anchor}} \sim \mathcal{N}(\mathbf{0}, \mathbf{I})$.
Each frame's latent states are obtained by interpolating between the shared anchor latent and a Gaussian random variable, $\mathbf{z}_T^f = \alpha \mathbf{z}_T^{\mathrm{anchor}} + (1-\alpha) \mathbf{z}^f$, where random variable $\mathbf{z}^f \sim \mathcal{N}(\mathbf{0}, \mathbf{I})$ and $\alpha$ is a hyperparameter to control the degree of correlation.

Similarly, CCEdit~\cite{rwy+23} adopts a trident-like architecture: an inflated T2I UNet with temporal layers and two separate hyper-networks~\cite{zra23,cxl+24} dedicated to structure and appearance. 
An anchor frame from the source video is edited into $\mathbf{z}_0^{\mathrm{anchor}}$ by the I2I method before being fed into the appearance hyper-network to extract pyramid features. 
These appearance features are subsequently added to the UNet~\cite{cxl+24} encoder blocks. This design facilitates the propagation of the edited anchor frame throughout the video. 
Similarly, the structural features are extracted by another hyper-network and incorporated into the decoder of the denoising UNet to provide structural guidance~\cite{zra23}.
For inference, the latent state for each frame $\mathbf{z}_T^f$ is independently initialized from $\mathcal{N}(\alpha \mathbf{z}_0^{\mathrm{anchor}}, \mathbf{I})$ rather than $\mathcal{N}(\mathbf{0}, \mathbf{I})$. This approach enables the anchor frame to contribute a portion of low-frequency information, adjusted by a small factor $\alpha$.

\vspace{6pt} \noindent$\bullet$~\textbf{Motion Conditions.}
Despite being equipped with temporal self-attention layers, preliminary experiments revealed a deficiency of temporal smoothness~\cite{lww+23}. 
MoCA~\cite{yba+23} and FlowVid~\cite{lww+23} integrate the edited first frame and optical flow~\cite{zj20} into the model input. Specifically, MoCA converts the optical flow into RGB frames and channel-wisely concatenates it with the edited first frame ${\mathbf{z}}^{1}_0$ and intermediate noisy latents $\mathbf{z}_t$. Conversely, FlowVid utilizes the source optical flow $\mathcal{F}^{1 \rightarrow f}$, the warpings of the edited first frame $\mathcal{F}^{1 \rightarrow f}({\mathbf{z}}^{1}_0)$, and occlusion masks $\mathcal{O}^{1 \rightarrow f}$. 
Moreover, FlowVid integrates a ControlNet-like architecture~\cite{zra23} with depth maps~\cite{RanftlLHSK22} and optical flows~\cite{zj20} conditions.

These motion conditions can also be integrated in a training-free manner. For instance, VideoControlNet~\cite{zd23}, proposed by Hu \& Xu, enables video editing by leveraging T2I diffusion models~\cite{zra23} and optical flow~\cite{zj20} without requiring parameter fine-tuning.
The first frame is edited using ControlNet~\cite{zra23} with Canny or depth controls. Then it divides subsequent frames into groups of size $g$, where the last frame of each group is generated by warping the previous edited frame with dense optical flow $\mathcal{F}^{f \rightarrow f+g}({\mathbf{x}}^{f})$ and refining the occlusions through pre-trained inpainting model~\cite{zra23}. Finally, the intermediate frames are generated by bidirectionally warping the adjacent edited frames.

\vspace{6pt} \noindent$\bullet$~\textbf{Compositional Conditions.}
Beyond providing a single condition, VideoComposer \cite{WangYZCWZSZZ23} enables simultaneous multi-conditional control by integrating diverse inputs such as sketch images \cite{0002LYH00P021}, depth maps \cite{RanftlLHSK22}, and motion vectors \cite{VadakitalDLTLR22}.
It employs Spatio-Temporal Condition (STC) encoders, consisting of spatial convolution layers followed by temporal transformer layers.
The multimodal conditions from STC encoders are aggregated through element-wise addition and concatenated with the noisy latent state $\mathbf{z}_t$ along the channel dimension, facilitating collaborative control.

\begin{figure}[tbp]
    \centering
    \includegraphics[width=\linewidth]{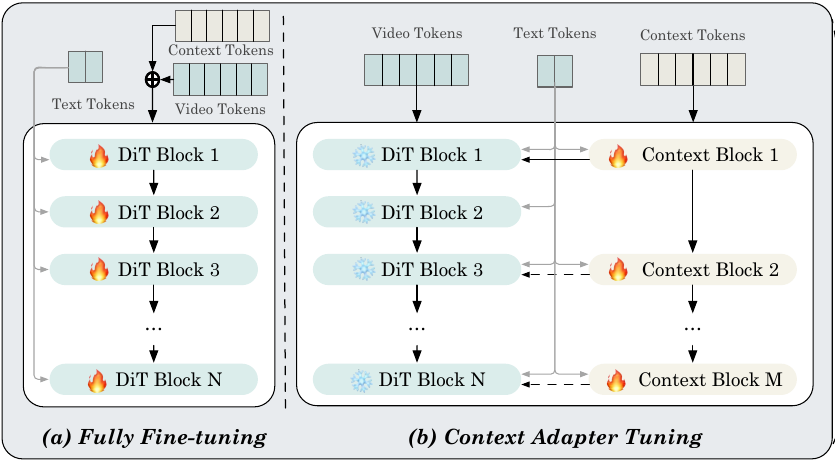}
    \caption{
    The architecture of VACE~\cite{vace}.
    The VACE \cite{vace} training strategies: (a) Fully Fine-tuning and (b) Context Adapter Tuning.
    Figure adapted from VACE~\cite{vace}.
    }
    \label{fig:vace}
\end{figure}

VACE \cite{vace} further extends compositional conditioning to include face, object, scribble, and layout modalities, enabling a universal, all-in-one video editing framework. 
These conditions are concatenated along the channel dimension with optional masks and processed through a trainable context embedder, analogous to the STC in VideoComposer \cite{WangYZCWZSZZ23}.
VACE introduces two strategies for integrating conditional contexts (\cref{fig:vace}). 
The first is a direct integration approach, where encoded conditional context tokens are added to the diffusion latents as input, followed by full fine-tuning of the video Diffusion Transformer (DiT) \cite{PeeblesX23}. 
Alternatively, to achieve faster convergence, VACE employs Res-Tuning \cite{res-tune} (\cref{fig:vace}, b). In this configuration, the DiT blocks remain frozen, while auxiliary context blocks, initialized from the DiT architecture, process the conditional inputs. 
The outputs of these context blocks are then integrated into the main DiT branch as additive signals, effectively guiding the model during generation and editing tasks.
This progression reflects a boarder trend toward leveraging increasingly complex conditioning in video editing.

\subsubsection{Training Modification}
Supervision plays a crucial role in tuning-based methods.
Various supervision training proposed specifically for video editing has demonstrated good results.

\vspace{6pt} \noindent$\bullet$~\textbf{Motion-Oriented Loss for Test-Time Tuning.}
MotionDirector~\cite{ryj+23} decouples the appearance and motion learning as AnimateDiff~\cite{yca+23}. Specifically, it first optimizes the spatial layers using LoRAs~\cite{eyp+22}. Then, it combines the trained LoRAs from the first stage and optimizes the LoRAs of the temporal attention blocks. To mitigate appearance bias, it introduces a residual operation:
\begin{equation}
    \delta(\epsilon^f) = \sqrt{1+\beta^2} \epsilon^f - \beta \epsilon^{\mathrm{anchor}}
\end{equation}
where the $\epsilon^{\mathrm{anchor}}$ is randomly selected from $\{\epsilon^f\}_{f=1}^{F}$ during training and $\beta$ serves as the decentralization factor. The residual operation is used to formulate a debiased temporal loss by substituting the $\epsilon$ from \cref{eq:dm-loss} into $\delta(\epsilon)$:
\begin{equation}
    \label{eq:motion-director-loss}
    \mathcal{L}_{\mathrm{debiased}} = \mathbb{E}\big[ \|
    \delta(\epsilon_t) - \delta\big(\epsilon_\theta (
        \mathbf{z}_t, t, y
    ) \big)
    \|_2^2 \big]
\end{equation}
During the optimization stage of temporal LoRAs, MotionDirector combines \cref{eq:dm-loss,eq:motion-director-loss} $\mathcal{L}_{\mathrm{MotionDirector}}=\mathcal{L}_{\mathrm{simple}} + \mathcal{L}_{\mathrm{debiased}}$ as the final loss function. 
Similarly motivated, VMC~\cite{hgj23} defines a residual operation with a pre-defined stride: $\delta(\epsilon^f) = \epsilon^{f+\mathrm{stride}} - \epsilon^{f}$.
Their empirical design opts for cosine similarity
$S_{\cos}(\mathbf{a}, \mathbf{b}) = 1 - {\langle \frac{\mathbf{a}}{\| \mathbf{a} \|_2}, \frac{\mathbf{b}}{\| \mathbf{b} \|_2} \rangle}$ over the Euclidean distance to optimize the temporal attention layers:
\begin{equation}
    \label{eq:vmc-loss}
    \mathcal{L}_{\mathrm{VMC}} = \mathbb{E}\big[
    S_{\cos}\big(
        \delta(\epsilon_t), 
        \delta(\epsilon_\theta (\mathbf{z}_t, t, y)) 
    \big)
    \big]
\end{equation}

\vspace{6pt} \noindent$\bullet$~\textbf{Textural Embedding Fine-tuning.}
Many approaches choose to finetune the feature embeddings that directly reflect semantic information.
SAVE~\cite{ywj+23} expands upon the concept of textural inversion~\cite{ryy+23,nyv+23} by introducing separate sets of token embeddings for appearance and motion. 
The optimization begins by learning the token embeddings representing appearance information. Subsequently, SAVE focuses on optimizing the time-variant motion token embeddings, facilitating a disentangled learning process: $\tau_{\mathrm{motion}} = \mathbf{W} \cdot \mathcal{P}(\tau_{\mathrm{motion}})$, where $\tau_{\mathrm{motion}} \in \mathbb{R}^{d_{\tau}\times F\times L}$ represents the expanded motion embedding.
It is initialized by duplicating the original tokens along the temporal dimension to achieve rapid coverage. $\mathcal{P}(\cdot)$ represents temporal positional encoding.
While optimizing the motion tokens, SAVE employs a cross-attention map-based regulation to prevent overspreading~\cite{cwq+23}.

\vspace{6pt} \noindent$\bullet$~\textbf{Image-Video Mixed Fine-tuning.}
Dreamix~\cite{eed+23} fine-tunes a pre-trained T2V model, Imagen Video~\cite{jwc+22}, using a combination of frames and video segments extracted from the source video. This hybrid training approach ensures the preservation of both structure and motion characteristics.
Besides, Dreamix adopts the SDEdit~\cite{cyy+22} schema, which initiates the sampling process by adding noise to the source video to perturbate its high-frequency structures.

\vspace{6pt} \noindent$\bullet$~\textbf{Instruction-Based Fine-tuning.}
An inherent limitation of description-based methods is requiring users to provide both source and target prompts, necessitating a text-video paired dataset format.
Beyond the increased workload, user descriptions of non-editing areas occasionally impact overall quality.
Pioneering works in image editing~\cite{bhe23} have introduced methods requiring only a single editing instruction, significantly enhancing efficiency. Researchers have extended this approach to video editing.

InstructVid2Vid~\cite{bjs+23}, proposed by Qin~\etal, formulates this task as a supervised learning problem and introduces a data curation pipeline. Firstly, BLIP ~\cite{0001LXH22} generates captions for the source videos. These captions are utilized to generate target captions and editing instructions using ChatGPT~\cite{chatgpt}. Subsequently, the source videos, source captions, and target captions are leveraged by off-the-shelf editing methods like Tune-A-Video~\cite{wgw+23} to produce the target video.
New triplet samples include editing instructions, source videos, and edited videos.
They are used to train the instruction-based video editing model. InstructVid2Vid also incorporates a delta-frame regularization $\mathcal{L}=\mathcal{L}_{\mathrm{simple}} + \alpha \sum_{f=2}^{F} (\epsilon^f - \epsilon^{f-1})$ to maintain consistency between consecutive frames.
During inference, the additional preprocessing steps, such as per-video fine-tuning or latent inversion, are no longer needed, thereby enhancing efficiency.

Instead of collecting videos, InsV2V~\cite{ZhangLNHGL23} initiates by constructing prompts~\cite{BainNVZ21,bhe23}. It employs a T2V model~\cite{jhd+23} and P2P~\cite{hmt+22} to generate the source and target videos from prompts. 
It focuses on editing long videos. Simply splitting a long video into non-overlapping chunks of size $F$ and processing each chunk individually results in visible boundaries between chunks.
InsV2V introduces a motion compensation approach to address this issue. It appends the last $a$ frames from the $(i-1)$-th chunk $\mathbf{z}_t^{f-a:f-1}(i-1)$ into the input of $i$-th chunk ${\mathbf{z}}_t^{f:f+F}(i)$ and predicts $[{\epsilon}_\theta^{f-a:f-1}(i-1), {\epsilon}_\theta^{f:f+F}(i)]$. Subsequently, the prediction discrepancies are corrected:
\begin{equation}
    {\epsilon}_\theta^{f:f+F}(i) \leftarrow {\epsilon}_\theta^{f:f+F}(i) + \frac{1}{a}\sum_{j=1}^{a}
    ({\epsilon}_\theta^{f-j}(i-1) -{\epsilon}_\theta^{f-j}(i))
\end{equation}

\vspace{6pt} \noindent$\bullet$~\textbf{Distillation.}
Emu Video Edit (EVE) \cite{uay+24} is built upon a T2I backbone model \cite{xjc+23}, along with separate image editing \cite{sau+23} and video generation \cite{rma+23} adapters. Directly integrating these adapters into the shared T2I backbone using a plug-and-play approach may lead to conflicts and artifacts.
To address these issues without requiring supervised video editing data, EVE introduces a factorized diffusion distillation method to transfer the knowledge from the adapters to a student video editing model.
Specifically, during each iteration, the student model first generates an edited video $\hat{\mathbf{z}}_0$, and the Score Distillation Sampling (SDS) \cite{PooleJBM23} is applied using the outputs from each teacher adapter:
\begin{equation}
    \mathcal{L}_{\mathrm{SDS}} = \mathbb{E}\big[
        c(t) \cdot \hat{\mathbf{z}}_0 \cdot 
        \mathrm{sg}(\epsilon - \epsilon_\theta^T(\hat{\mathbf{z}}_{t}, t))
    \big]
\end{equation}
where $c(t)$ is a weighting function, $\mathrm{sg}(\cdot)$ is the stop gradient operation, $epsilon$ is the sampled noise, $\epsilon_\theta^T(\cdot)$ is the teach model, and  $\hat{\mathbf{z}}_t$ represents the noised version of $\hat{\mathbf{z}}_0$, generated using $\epsilon$ and \cref{eq:ddpm-t-step}. In addition to the SDS loss derived from the two teacher models, EVE incorporates an adversarial loss \cite{ijm+14,adar+23} for each teacher to mitigate blurriness effects. These various losses are combined to train the student video editing model.

\subsection{Attention Feature Injection}\label{subsec:feat-inject}
The hidden features in diffusion models inherently capture semantic correspondence\cite{tjw+23}.
Analogous to image editing feature injection techniques (discussed in \cref{subsubsec:img-editing}), manipulating hidden features to edit a video is a promising direction.
We classify related methods into two categories based on the origination of the injected hidden features.

\subsubsection{Inversion-Based Feature Injection}
Inversion-based injection assumes that the reconstruction branch holds the ``ground truth'' for layout and appearance of the source videos.
Hidden features derived from the reconstruction branch can facilitate the generation of the target video.
The challenge here is the ``editability-fidelity trade-off'': too much injection prevents the prompt from taking effect, while too little leads to temporal flickering.
Recent multi-branch architectures~\cite{bhw+24,kwr+24} attempt to solve this by isolating motion cues from appearance cues across separate reference streams.

\vspace{6pt} \noindent$\bullet$~\textbf{Dual-Branch.}
Video-P2P~\cite{lzl+23} and Vid2Vid-Zero~\cite{wxl+23} expand the Prompt-to-Prompt (P2P)~\cite{hmt+22} framework into the realm of video editing. 
They both employ cross-attention map injection~\cite{hmt+22} and null-text inversion~\cite{mha+23}, as discussed in \cref{subsubsec:img-editing}. The null-text embeddings are shared across all frames to enhance semantic consistency and reduce memory usage.
Video-P2P replaces spatial self-attention layers with sparse-causal attention, introduces temporal attention layers, and implements one-shot tuning to learn temporal consistency~\cite{wgw+23}.
Vid2Vid-Zero, on the other hand, highlights the necessity of global spatio-temporal attention layers, with key and value features sourced from all frames. Even without parameter tuning, its inflated 2D layers empirically demonstrate effective temporal modeling.

Fate-Zero~\cite{qcz+23} 
injects both cross-attention and self-attention maps, reaping the benefits from P2P~\cite{hmt+22} and Plug-and-Play (PnP)~\cite{tgb+23}. The fusion masks for self-attention injection are generated by binarizing the cross-attention maps~\cite{afl23,alf22}.
Edit-A-Video~\cite{skl+24} has a similar attention map injection schema as Fate-Zero and uses null-text inversion~\cite{mha+23} to mitigate errors associated with latent inversion and classifier-free guidance~\cite{hs22}. Additionally, it interpolates the fusion masks between the first frame and the current frame, weighted by self-attention maps, thereby ensuring temporal smoothness in the edited area.

Make-A-Protagonist~\cite{zxh+23} follows the path of PnP~\cite{tgb+23} and MasaCtrl~\cite{cwq+23}, injecting the latent features in residual blocks and self-attention maps from the reconstruction branch into the editing branch.
Rather than deriving the mask from attention maps as in prior works~\cite{cwq+23}, it utilizes Grounded SAM~\cite{rlz+24} to obtain the first frame segmentation mask and Xmem~\cite{cs22} to propagate it for precise fusion.

\vspace{6pt} \noindent$\bullet$~\textbf{Multiple Branches.}
UniEdit~\cite{bhw+24} and AnyV2V~\cite{kwr+24} disentangle appearance and motion injection by introducing two auxiliary branches: a reconstruction and a motion-reference branch. All branches are initialized by pre-trained video generation models~\cite{wcm+23,zwz+23,ryz+24,cwz+23}.
The reconstruction branch injects spatial self-attention query and key features at earlier sampling steps to maintain structural consistency. Furthermore, spatial self-attention value features from the reconstruction branch are selectively injected to preserve non-edited regions. 
The motion-reference branch contributes to the desired motion by injecting only its temporal self-attention query and key features, thereby avoiding potential structural instability.
Although sharing the injection workflow, UniEdit and AnyV2V diverge in the base models.
UniEdit leverages a T2V model, LaVie~\cite{wcm+23}, and efficiently incorporates text prompts.
AnyV2V utilizes I2I experts~\cite{geb15,wbw+24,chl+23,bhe23} to edit the first frame, and subsequently applies a video editing pipeline based on pre-trained I2V models~\cite{zwz+23,ryz+24,cwz+23} to enhance compatibility across a wide range of editing tasks.

\subsubsection{Motion-Based Feature Injection}
In contrast to inversion methods that rely on latent reconstruction, motion-based injection leverages explicit geometric correspondence (\eg, optical flow or token similarity).
The intrinsic motion information within the source video provides valuable prior knowledge for indicating temporal correspondence and guiding inter-frame feature injection.
This transition reflects a shift from relying on the model’s internal hidden understanding to enforcing consistency through external pixel-level trajectories.

TokenFlow~\cite{mos+23} uses inter-frame feature similarity.
It first processes several sampled keyframes by replacing spatial self-attention with a global spatio-temporal self-attention mechanism. 
TokenFlow then identifies the nearest neighbor from adjacent keyframes for each token.
For a token at frame $f$ with spatial coordinates $(x_f, y_f)$, its nearest neighbor tokens at adjacent past keyframe $f-$ and adjacent future keyframe $f+$ are denoted as $(x_{f-}, y_{f-})$ and $(x_{f+}, y_{f+})$, respectively. Subsequently, each frame is independently denoised with inter-frame feature injection:
\begin{equation}
    \mathbf{h}^f_{[x_f, y_f]} \leftarrow  
        \alpha_f\mathbf{h}^{f-}_{[x_{f-}, y_{f-}]} + 
        (1-\alpha_f)\mathbf{h}^{f+}_{[x_{f+}, y_{f+}]}
\end{equation}
where $\alpha_f \in (0, 1)$ is a scalar proportional to the distance between frame $f$ and its adjacent keyframes $j\pm$. And $\mathbf{h}$ denotes the self-attention hidden features.

In contrast to the feature similarity-based approach in TokenFlow, FLATTEN~\cite{ymc+23} selects injected features based on optical flow.
Specifically, for a token ${\mathbf{x}^s}^1_{[x_1, y_1]}$ at the first frame of the source video ${\mathbf{x^s}}^1$ with spatial coordinates $(x_1, y_1)$, optical flow estimation methods~\cite{LaiHWSYY18} can track its spatial correspondence across the remaining $F-1$ frames, denoting it as a trajectory: 
\begin{equation}
\mathcal{T}({\mathbf{x^s}}, x_1, y_1) = \{{\mathbf{x^s}}^1_{[x_1, y_1]}, \cdots, {\mathbf{x^s}}^F_{[x_F, y_F]}\}
\end{equation}
During editing, its temporal-guided attention mechanism pairs a query token with key and value tokens from the same trajectory in other frames:
\begin{align*}
    \mathbf{Q}^f_{[x_f, y_f]} &= \mathbf{h}^f_{[x_f, y_f]} \numberthis \label{eq:flatten}\\
    \mathbf{K}^f_{[x_f, y_f]} &= \mathbf{V}^f_{[x_f, y_f]} = \mathcal{T}(\mathbf{h}, x_1, y_1) \setminus \{\mathbf{h}^f_{[x_f, y_f]}\}
\end{align*}
It is important to note that temporal-guided attention does not involve positional encoding, projection matrices, or feed-forward layers, making seamless integration without additional training.

FRESCO~\cite{syz+24} follows a similar approach to FLATTEN, with additional steps to refine the value features.
It performs one forward and reverse DDPM~\cite{jap20} step on a source frame ${\mathbf{z}^s}^f$, and extracts its query features ${\mathbf{Q}^s}^f \in \mathbb{R}^{d \times N}$ and key features ${\mathbf{K}^s}^f\in \mathbb{R}^{d \times N}$. 
FRESCO then introduces a spatial-guided attention mechanism, utilizing these features to reweight the query features in the editing video:
\begin{equation}
    \mathbf{Q}^f \leftarrow \mathrm{Softmax}(({\mathbf{Q}^s}^{f})^{T} \cdot {\mathbf{K}^s}^{f}/{s \sqrt{d}}) \cdot \mathbf{Q}^f
\end{equation}
where $s$ is a scaling factor. 
Subsequently, it utilizes the efficient cross-frame attention to enhance the value features with cross-frame information:
\begin{equation}
    \mathbf{V}^f \leftarrow \mathrm{Softmax}\big(
    (\mathbf{Q}^{f})^{T} \cdot \mathbf{K}_{[\mathbf{p}]}/{\sqrt{d}} \big)
    \cdot \mathbf{V}_{[\mathbf{p}]}
\end{equation}
Here, $\mathbf{p}$ denotes a set of attended coordinates: 
all coordinates in the first frame are included, and subsequent frames only include previously non-observed areas.
The refined value features are utilized in a temporal-guided attention mechanism, similar to FLATTEN~\cite{ymc+23} in \cref{eq:flatten}.
Moreover, it finetunes the input features of each decoder layer. The source and edited video undergo one DDPM~\cite{jap20} forward and inverse step. 
The input features at the $f$-th frame of source and target video in the inverse process are denoted as $\mathbf{h^s}^f \in \mathbb{R}^{d \times N}$ and ${\mathbf{h}^f} \in \mathbb{R}^{d \times N}$, respectively. These feature embeddings are optimized by minimizing a spatio-temporal alignment loss:
\begin{align*}
    & \mathcal{L} = \mathcal{L}_{\mathrm{temporal}} + \lambda \mathcal{L}_{\mathrm{spatial}} \numberthis\\
    & \mathcal{L}_{\mathrm{temporal}} = \sum_{f=1}^{F-1}\| 
        \mathcal{O}^{f\rightarrow f+1}(\mathbf{h}^{f+1} - 
        \mathcal{F}^{f\rightarrow f+1}(\mathbf{h}^f)) 
    \|_1\\
    & \mathcal{L}_{\mathrm{spatial}} = \lambda\sum_{f=1}^{F}\|
        {\mathbf{h^s}^f}^T \cdot {\mathbf{h^s}^f} - 
        {\mathbf{h}^f}^T \cdot {\mathbf{h}^f}
    \|_2^2
\end{align*}
where $\lambda$ is a scaler used to strike the balance.

\subsection{Diffusion Latent Manipulation}\label{subsec:diffusion-steps}
It is commonly hypothesized that the diffusion process operates independently of other design components~\cite{kaa+22}, such as network architecture and training configuration. From the perspective of the diffusion process, the noise predictor can be treated as a black box~\cite{whn+21,wch+21,kaa+22}. This section reviews diffusion latent states initialization and transition technologies for video editing.

\subsubsection{Latent Initialization}
\begin{figure}[t]
    \centering
    \includegraphics[width=\linewidth]{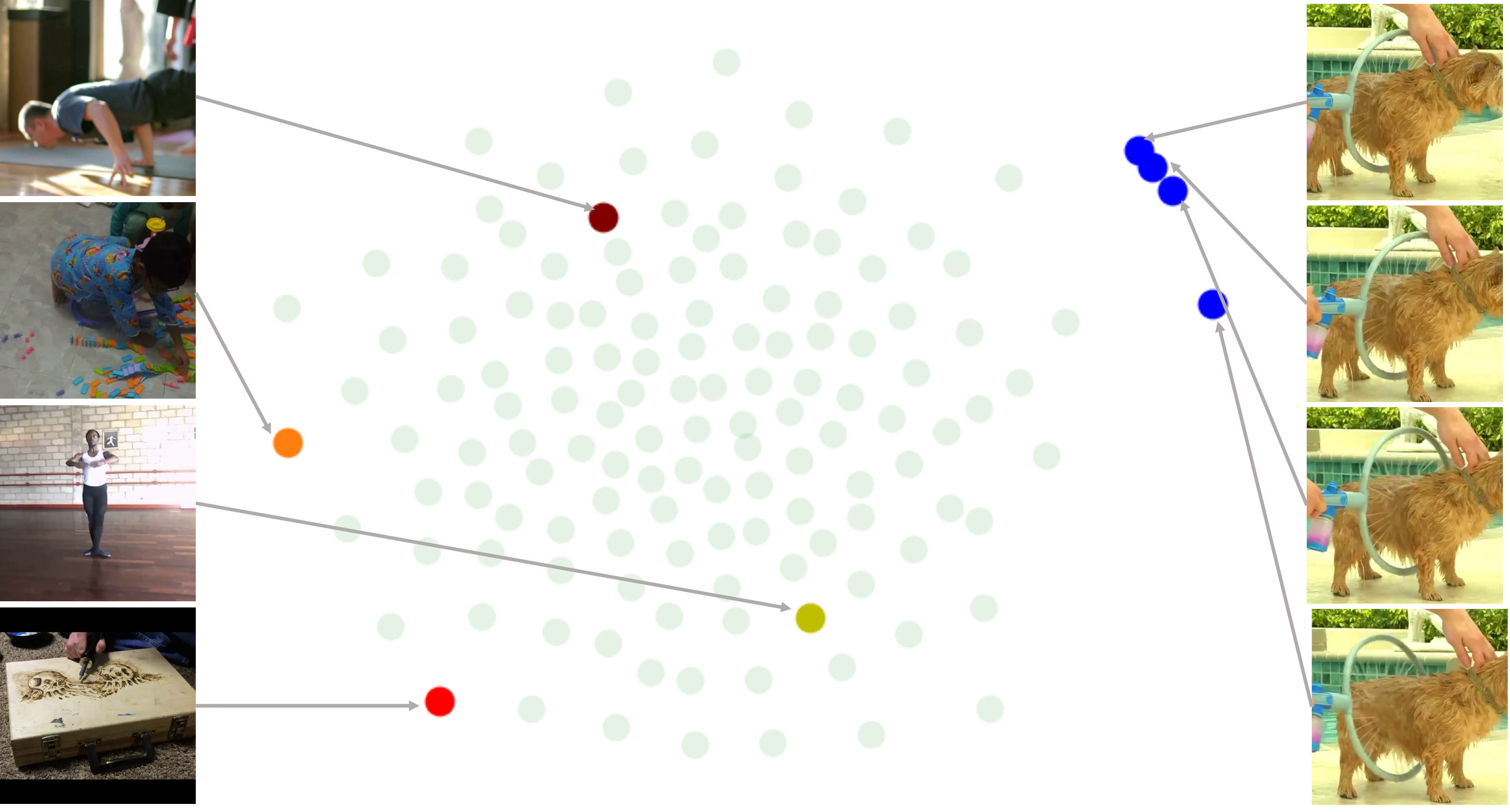}
    \caption{The t-SNE visualization illustrates the distribution of $\mathbf{z}_T$. 
    The samples are obtained from latent inversion on real video frames using a pre-trained T2I diffusion model. The green dots in the background represent samples drawn from \iid Gaussian distribution for reference. Blue dots represent successive frames from the same video, clustered closely together. While frames from different videos (yellow and red dots) are scattered and exhibit no correlation. The plot is adapted from PYoCo~\cite{ssg+23}.
    }
    \label{fig:pyoco}
\end{figure}

As shown in \cref{fig:pyoco},  the initial latent states $\mathbf{z}_T$ from consecutive video frames demonstrate significant correlation. Simply extending the image latent state initialization to video results in suboptimal performance~\cite{wjp+23}.
While latent inversion methods~\cite{sme20,mha+23} provide a solution, they require additional sampling steps for each source video.
On the other hand, methods such as LAMP~\cite{rlt+23} and CCEdit~\cite{rwy+23} employ simple initialization strategies to achieve satisfactory performance, underscoring the significance of latent state initialization in video diffusion models. 

Chen's Control-A-Video~\cite{wjp+23} proposes a residual-based latent state propagation approach:
\begin{equation}
    \label{eq:cav}
    \mathbf{z}_T^f = \alpha^f \mathbf{z}_T^{f-1} + (1 - \alpha^f) \mathbf{\nu}
\end{equation}
where $\alpha^f = | \mathbf{x}^f - \mathbf{x}^{f-1} | < \mathcal{R}$ serves as the mask to ensure that unchanged areas exhibit the same initialization, $\mathcal{R}$ represents a pre-defined threshold, and $\mathbf{\nu} \sim \mathcal{N}(\mathbf{0}, \mathbf{I})$ is a standard Gaussian random variable. Similarly motivated, the initialization of Ground-A-Video~\cite{hj23} is also formulated as \cref{eq:cav} with a little difference: $\alpha^f = | \mathcal{F}^{f-1 \rightarrow f} | < \mathcal{R}$.
Chen~\etal~\cite{wjp+23} also introduces a flow-based diffusion latent state initialization scheme:
\begin{equation}
    \mathbf{z}_T^f = \mathcal{F}^{f-1 \rightarrow f}(\mathbf{z}_T^{f-1})
\end{equation}
where the optical flow operation $\mathcal{F}$ is defined in \cref{subsec:optical-flow}.

Text2Video-Zero~\cite{KhachatryanMTHW23} constructs the inter-frame linkage of latent states in an earlier diffusion step $T'=T-\Delta T$, in contrast to Chen~\etal~\cite{wjp+23}, which does so at step $T$.
It begins by randomly initializing the first frame $\mathbf{z}_T^1 \sim \mathcal{N}(0, I)$, followed by $\Delta T$ denoising steps to obtain $\mathbf{z}_{T'}^1$. 
Text2Video-Zero subsequently defines a sequence of translation vectors for each frame $\delta^f = (\delta^f_x, \delta^f_y) \in \mathbb{R}^2$. These translation vectors are employed to warp the first frame $\mathbf{z}_T'^1$ into the subsequent frames $\mathbf{z}_{T'[i, j]}^f \leftarrow \mathbf{z}_{T'[i + \delta^f_x, j + \delta^f_y]}^1$.
This warping can be optionally disabled in specific areas indicated by off-the-shelf object detection bounding boxes~\cite{WangLFSLY22}, preserving the consistency of the background.

Rather than manual construction, Chu's VideoControlNet~\cite{esj+23} optimizes the inter-frame correlation with optical flow supervision:
\begin{equation}
    \mathcal{L} = \sum_{f=1}^F
    \Big(\frac{1}{f}\sum_{a=1}^{f} 
    \mathcal{O}^{a \rightarrow f} \odot \big(
        \epsilon_\theta(\mathbf{z}_T^f) - \epsilon_\theta(\mathcal{F}^{a \rightarrow f}(\mathbf{z}_T^a))
    \big)\Big)
\end{equation}
where $\epsilon_\theta(\cdot)$ is a pre-trained T2I ControlNet~\cite{zra23}, $\odot$ represents the element-wise multiplication.
Chu~\etal also suggests first optimizing keyframes and then interpolating the remaining frames to accelerate the optimization.

\subsubsection{Latent Transition}
Enhancing the reverse latent transition (denoising) process aids the convergence of $\{\mathbf{z}_t\}_{t=T}^0$ towards a temporally consistent video distribution.
Pix2Video~\cite{CeylanHM23} demonstrates inter-frame smoothness via gradient descent. Rather than the routine reverse diffusion sampling step as described in \cref{eq:ddim}, it updates the latent states by:
\begin{equation}
    \mathbf{z}^f_{t-1} \leftarrow \mathbf{z}^f_{t-1} - 
    \lambda_{t} \nabla_{\mathbf{z}^{f}_t}
    \|\mathbf{z}^{f-1}_{t\rightarrow 0} - \mathbf{z}^{f}_{t\rightarrow 0} \|_2
\end{equation}
where $\lambda_t$ represents the step size, and $\mathbf{z}^f_{t\rightarrow 0}$ denotes the estimated $\mathbf{z}^{f}_0$, consistent with \cref{eq:ddim}. This guidance prevents temporal inconsistency at each denoising step.

ControlVideo~\cite{yyd+23}, proposed by Zhang~\etal, enforces the smoothness of the $\mathbf{z}_{t\rightarrow 0}$.
Each middle frame is interpolated using its two adjacent frames:
\begin{equation}
    \mathbf{z}_{t\rightarrow 0}^f \leftarrow
        \mathcal{E}\Big(\mathrm{Interpolate}\big(
            \mathcal{D}(\mathbf{z}_{t\rightarrow 0}^{f-1}),
            \mathcal{D}(\mathbf{z}_{t\rightarrow 0}^{f+1})
        \big)\Big)
\end{equation}
Here, the interpolated frames $f \in \{f_0, f_0+2, \cdots \}$ are sampled in an interleaved manner, with $f_0 \in \{0, 1\}$. The functions $\mathcal{E}(\cdot)$ and $\mathcal{D}(\cdot)$ represent the VAE encoding and decoding.
Subsequently, \cref{eq:ddim} is utilized to update the diffusion states. After each denoising step, the interpolation frames are updated with $f_0 \leftarrow 1 - f_0$.
The sequential combination of two steps effectively smooths all frames.

Rerender-A-Video~\cite{YangZLL23} exhibits inter-frame fusion in VAE latent and pixel space, ensuring simultaneous consistency in semantics and texture. The latent fusion is achieved by updating $\mathbf{z}_{t \rightarrow 0}$ using the warped first frame $\mathcal{F}^{1\rightarrow f}(\mathbf{z}^1_{t \rightarrow 0})$ and its corresponding occlusions $\mathcal{O}^{1\rightarrow f}$:
\begin{equation}
    \label{eq:rerender-latent}
    \mathbf{z}^f_{t \rightarrow 0} \leftarrow 
        \mathcal{O}^{1\rightarrow f} \mathbf{z}^f_{t \rightarrow 0} +
        (1 - \mathcal{O}^{1\rightarrow f}) \mathcal{F}^{1\rightarrow f}(
            \mathbf{z}^1_{t \rightarrow 0})
\end{equation}
where the optical flow is estimated from the source video.
The fusion in the VAE latent space only occurs during the early denoising steps. 
To enhance low-level feature consistency, the first-round denoising output $\hat{\mathbf{x}}$ is successively fused with the previous frame and the first frame:
\begin{align*}
    \hat{\mathbf{x}}^f & \leftarrow
        \mathcal{O}^{f-1\rightarrow f} \hat{\mathbf{x}}^f +
        (1 - \mathcal{O}^{f-1\rightarrow f}) \mathcal{F}^{f-1\rightarrow f}(
            \hat{\mathbf{x}}^{f-1})\\
    \hat{\mathbf{x}}^f & \leftarrow
        \mathcal{O}^{1\rightarrow f} \hat{\mathbf{x}}^f +
        (1 - \mathcal{O}^{1\rightarrow f}) \mathcal{F}^{1\rightarrow f}(
            \hat{\mathbf{x}}^{1}) \numberthis
\end{align*}
Within the occlusion mask area $\mathcal{O}^f = \mathcal{O}^{1\rightarrow f} \cap \mathcal{O}^{f-1\rightarrow f}$, the frame $\hat{\mathbf{x}}^f$ remains unaltered. Subsequently, Rerender-A-Video regenerates the edited video with the guidance of $\hat{\mathbf{x}}$ at each denoising step:
\begin{equation}
    \label{eq:rerender-pix}
    \hat{\mathbf{z}}_{t-1}^f \leftarrow \mathcal{O}^f \mathbf{z}_{t-1}^f +
        (1 - \mathcal{O}^f) \hat{\mathbf{z}}_{t-1}
\end{equation}
where $\hat{\mathbf{z}}_{t-1}$ is obtained by applying \cref{eq:ddpm-t-step} on $\hat{\mathbf{z}}_0 = \mathcal{E}(\hat{\mathbf{x}})$. Note that latent space fusion (shown in \cref{eq:rerender-latent}) occurs before \cref{eq:ddim}, while the second-round pixel space fusion (shown in \cref{eq:rerender-pix}) occurs after it.

Yatim \etal \cite{dro+23} adjusts the intermediate latent states before the latent transition step to align the motion and scene layout of the source video. They first introduce a feature descriptor dubbed Spatial Marginal Mean (SMM) to capture the semantic information:
\begin{equation}
    \mathrm{SMM}(f, \mathbf{z}_t) = \frac{1}{W \cdot 
 H}\sum_{i=1}^{W}\sum_{j=1}^{H} \mathbf{h}^f_{[i,j]}(\mathbf{z}_t)
\end{equation}
where $\mathbf{h}^f(\mathbf{z}_t) \in \mathbb{R}^{d \times W \times H }$ represents the DIFT features \cite{tjw+23} of the $f$-th frame, given the model input $\mathbf{z}_t$.
Instead of using the exact SMM, they use pairwise SMM difference to further remove appearance-related information:
\begin{equation}
    \delta(f_1, f_2, \mathbf{z}_t) = \mathrm{SMM}(f_1, \mathbf{z}_t) - \mathrm{SMM}(f_2, \mathbf{z}_t)
\end{equation}
The latent states are iteratively denoised with an additional adjustment step applied before the process described in \cref{eq:ddim}:
\begin{equation}
    \mathbf{z}_t \leftarrow \argmin_{\mathbf{z}_t} \sum_{f_1, f_2}
        \| \delta(f_1, f_2, \mathbf{z}_t) - \delta^s(f_1, f_2, \mathbf{z}^s_t) \|_2^2
\end{equation}
where $\delta^s(f_1, f_2, \mathbf{z}^s_t)$ denotes the pairwise SMM difference of the source video.

RAVE~\cite{obh+23} employs a `grid trick': video frames are divided into $g$ equal-sized groups. Frames within the same group are concatenated along spatial dimensions into a large grid image as the input of the pre-trained image-to-image diffusion model~\cite{zra23,tgb+23}. The frames within each grid maintain a certain level of consistency. After each denoising step, the frames are randomly permuted into different groups, enabling interaction with more frames. Over the $T$ denoising steps, each frame can interact with up to $T(g-1)$ other frames, promoting global consistency.
Such grid manipulation requires neither additional parameters nor extra training, resulting in an efficient video editing adaptation.

\subsection{Canonical Representation}\label{subsec:canonical}
\begin{figure}
\includegraphics[width=0.99\linewidth]{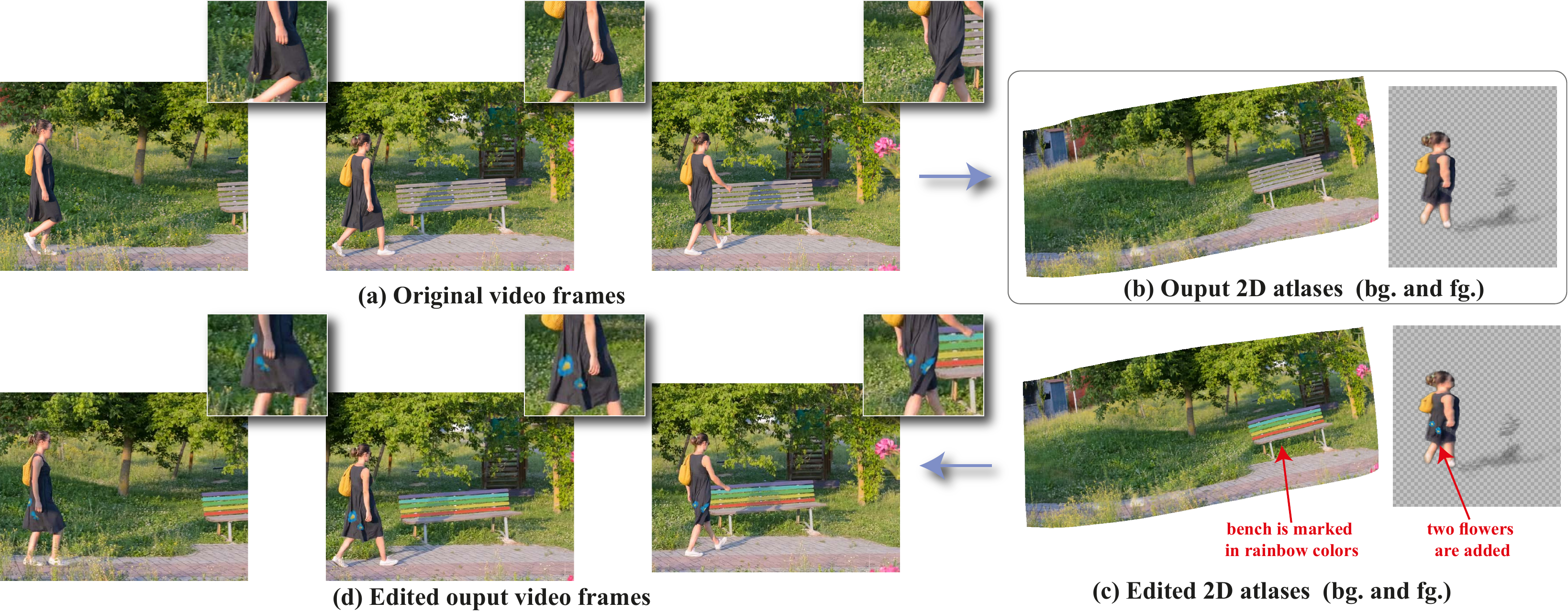}
\caption{
    Layered Neural Atlases (LNA)~\cite{KastenOWD21}.
    Original video frames are mapped to flattened 2D atlases by warping models. Edits on the 2D atlases can be propagated to all frames of the video. The plot is adapted from LNA~\cite{KastenOWD21}.
}
\label{fig:lna}
\end{figure}

The aforementioned methods represent the video clips as image frames. This section focuses on the video canonical representations, also known as atlases.
The canonical atlas can effectively handle the temporal axis of video data and bridge the gap between video and image modalities. This representation has inspired the adaptation of image editing methods for video applications.
More recently, as diffusion models have shown remarkable abilities in image editing~\cite{yjy+24}, researchers have integrated video canonical representations with diffusion models.
We begin with an introduction to video neural canonical representations, followed by introducing related video editing methods.

\subsubsection{Atlas Learning}
As shown in \cref{fig:lna}, the seminal work Layered Neural Atlases (LNA)~\cite{KastenOWD21} encodes a video pixel $\mathbf{x}^f_{[x, y]} \in \mathbb{R}$ to an atlas pixel $\mathbf{a}_{[u, v]} \in \mathbb{R}$ by learning the transformation from video voxel to atlas coordinates $(u, v) = \mathcal{W}(x, y, f) \in \mathbb{R}^2$. $\mathcal{W}(\cdot)$ represents the warping network, indicating the many-to-one mapping from 3D voxels to 2D atlas coordinates.
Specifically, it considers two separate atlases, $\mathbf{a}^{\mathrm{FG}}$ and $\mathbf{a}^{\mathrm{BG}}$, and two warping networks, $\mathcal{W}^{\mathrm{FG}}(\cdot)$ and $\mathcal{W}^{\mathrm{BG}}(\cdot)$, for foreground and background objects, respectively.
To demonstrate self-supervised learning, LNA designs a reconstruction process using a color prediction network $\mathcal{X}(\cdot)$ and an opacity prediction network $\mathcal{A}(\cdot)$. The video reconstruction from the atlas is defined by:
\begin{equation}
    \hat{\mathbf{x}}^f_{[x, y]} = 
        \alpha \mathcal{X}(u^{\mathrm{FG}}, v^{\mathrm{FG}}) 
        + (1-\alpha) \mathcal{X}(u^{\mathrm{BG}}, v^{\mathrm{BG}})
\end{equation}
where $\mathcal{X}(u, v)$ represents the predicted RGB value at atlas coordinate $(u, v)$ and $\alpha = \mathcal{A}(x, y, f)\in [0, 1]$ represents the blending factor of the predicted colors at voxel $(x, y, f)$.
During training, the warping network $\mathcal{W}(\cdot)$, the color prediction network $\mathcal{X}(\cdot)$, and the opacity prediction network $\mathcal{A}(\cdot)$ are trained together on the source video using a reconstruction loss with rigidity regularizations to maintain. 
Once the video is registered through training, edits can be applied directly to the canonical atlas.
The learned 2D-to-3D coordinates mapping $\mathcal{W}(\cdot)$ acts as the translation mechanism between the atlas and the video frames. The color prediction network $\mathcal{X}(\cdot)$ is not involved during the editing stage.

Although the effective representation of LNA can guarantee temporal consistency by design, it requires significant preprocessing time for each video, generally taking hours of training. CoDeF~\cite{hqy+23} addresses this by proposing an effective hash-based architecture for warping and reconstruction. 
It uses pre-trained SAM-tracking~\cite{yly+23} to derive the blending $\alpha$ values rather than learning them to reduce complexity. CoDeF also distills trajectory information from optical flow~\cite{zj20} to enhance consistency. The resulting atlas serves as an effective video canonical representation.

\subsubsection{Atlas Editing}
Editing the canonical atlases and then propagating the edits to the entire video offers the convenience of maintaining temporal consistency. However, the contents in the canonical atlas diverge from the distribution of natural images. Directly applying image editing methods trained on natural images may lead to unexpected deteriorated regions or semantic deformations~\cite{pcj+23}.

VidEdit~\cite{pcj+23} focuses on editing foreground objects using extracted structural information to guide the editing process. It applies the semantic segmentation method~\cite{ChengMSKG22} to delineate the region of interest during editing precisely. Additionally, it employs SDEdit~\cite{cyy+22} with HED~\cite{XieT15} conditions to provide critical structural guidance, thereby avoiding the pitfall of atlas distortion.
On the other hand, DiffusionAtlas~\cite{sht23} fine-tunes a pre-trained T2I diffusion model~\cite{rad+22} to adapt the atlas distribution. Atlas editing is demonstrated by generating editing prompts starting from the DDIM inverted latent states~\cite{sme20}. Additionally, DiffusionAtlas trains the foreground warping network $\mathcal{W}^{\mathrm{FG}}(\cdot)$ and the opacity prediction network $\mathcal{A}(\cdot)$ on the edited atlases to map the edited atlas back to video frames.

\subsubsection{Edit Propagation}
Another promising solution to address the issue of unnatural atlas distortion is to edit video frames and propagate these edits throughout the video using canonical atlases.
StableVideo~\cite{ChaiGWL23} starts by editing several keyframes. Specifically, the first keyframe is edited with structural conditions~\cite{zra23} and then warped into a partially edited atlas. It is worth noting that the atlas pixels not present in the first keyframe remain unmodified. The subsequent keyframes are initialized using their previous partially edited atlases and then undergo the SDEdit~\cite{cyy+22} to complete editing. Finally, an aggregation network trained on the source video frames and atlases is applied to combine these partially edited atlases into a fully edited atlas.

One challenge of using the pre-trained warping network $\mathcal{W}(\cdot)$ is shape editing.
Lee~\etal~\cite{LeeJCQ023} proposes propagating shape correspondences between source and target objects through the canonical atlas to achieve shape-aware video editing. They edit a keyframe using StableDiffusion~\cite{rad+22}.
Lee~\etal estimates semantic correspondences~\cite{TruongDYG22} and represents them as a set of deformation vectors.
Initially, the edited object is deformed to match the shape of its source object and propagates throughout the entire video. Simultaneously, the deformation vectors propagate throughout the video, bridging through the canonical space. These deformation vectors then restore the edited object shapes. Finally, it fine-tunes the foreground atlas network to enhance the visual quality of the edited frames.

\subsection{Novel Conditioning}\label{subsec:control}
In addition to text prompts, the newly proposed handle point prompts enhance the controllability of interactive video editing.
Furthermore, we also cover domain-specific methods that elucidate human action videos and leverage pose conditions in this section.

\subsubsection{Point Conditioning}\label{subsec:drag}
The expressiveness of text prompts is inherently limited by language modality.
Can editing be controlled by clicking on a frame to move, add, remove, or replace objects? Interactive point-based editing has emerged as a novel conditioning method that provides flexible, precise, and generalized control~\cite{PanTLLMT23}. Users can select handle and target points to move, add, or remove them. Edits applied to a single frame should propagate seamlessly across all frames.

\vspace{6pt} \noindent$\bullet$~\textbf{DragGAN and DragDiffusion.}
The seminal point-based image editing work, DragGAN~\cite{PanTLLMT23}, operates on the StyleGAN2~\cite{KarrasLAHLA20} hidden features $\mathbf{h} \in \mathbb{R}^{d\times W \times H}$ and includes two stages: motion supervision to enforce handle points $\{\mathbf{p}_i\}$ to move towards target points $\{\mathbf{t}_i\}$ and point tracking to update the positions of the handle points $\{\mathbf{p}_i\}$ after each movement step. 
In addition to the handle points, the motion supervision loss drives a surrounding patch $\Omega(\mathbf{p}_i, r_1) = \{\mathbf{q}: \|\mathbf{q} - \mathbf{p}_i \|\le r_1 \}$ to move toward target points, taking the form:
\begin{equation}
    \mathcal{L}_{\mathrm{motion}} = 
        \sum_i \sum_{\mathbf{q}_i \in \Omega(\mathbf{p}_i, r_1)}\|  \mathbf{h}_{[\mathbf{q}_i + \mathbf{d}_i]} - 
            \mathrm{sg}(\mathbf{h}_{[\mathbf{q}_i]}) \|_1
\end{equation}
where $\mathbf{d}_i = (\mathbf{t}_i - \mathbf{p}_i) / \| \mathbf{t}_i - \mathbf{p}_i \|_2 \in [0, 1]^{2}$ is an unit vector pointing from $\mathbf{p}_i$ to $\mathbf{t}_i$, and $\mathrm{sg}(\cdot)$ is the stop gradient operation.
Moreover, DragGAN allows users to provide a binary mask $\mathbf{M}\in \{0, 1
\}^{d\times W \times H}$ to prevent movement in the masked areas. It introduces a reconstruction loss:
\begin{equation}
    \mathcal{L}_{\mathrm{reconstruction}} = \| (\mathbf{h} - \mathbf{h}_0) \odot (\mathbf{1}-\mathbf{M}) \|_1
\end{equation}
where $\mathbf{h}_0\in \mathbb{R}^{d\times W \times H}$ denotes the initial hidden feature and $\odot$ denotes the element-wise multiplication.
After each optimization step, the position of handle points in the new hidden feature is unclear. The point tracking updates their position by searching over all neighbors around $\mathbf{p}_i$, formulated as:
\begin{equation}
    \mathbf{p}_i \leftarrow \argmin_{\mathbf{q}_i \in \Omega(\mathbf{p}_i, r_2)} \|
        \mathbf{h}_{[\mathbf{q}_i]} - {\mathbf{h}_0}_{[\mathbf{p}_i]}\|_1
\end{equation}
The optimization is iteratively applied to the new handle points until they reach the target points.

DragDiffusion~\cite{ycj+23} extends point-based image editing to diffusion models. It first performs sample-specific fine-tuning on the source image using LoRA~\cite{eyp+22}. Instead of operating on hidden features, it applies motion supervision to the latent states $\mathbf{z}_t$ with the initial hidden state obtained from the latent inversion of the source image as described in \cref{eq:ddim_inv}. As diffusion models exhibit greater generation diversity, DragDiffusion involves feature injection to retain attributes of the source image by injecting keys and values of the transformer blocks, as in MasaCtrl~\cite{cwq+23}.
Concurrently, DragonDiffusion~\cite{cxj+23} provides training-free interactive manipulation using energy functions and a visual cross-attention design as alternative solutions. FreeDrag~\cite{ling2024freedrag} enhances robustness through adaptive latent state updating and line search with backtracking.

\vspace{6pt} \noindent$\bullet$~\textbf{Point-Based Video Editing.}
DragVideo~\cite{yry+23} follows the DragDiffusion paradigm~\cite{ycj+23} and utilizes AnimateDiff~\cite{yca+23} as a pre-trained video backbone. It requires user points and masks on the first and last frames. 
To propagate these conditions across frames, It employs Persistent Independent Particles (PIPs)~\cite{bian2023context} for the handle points, linear interpolation for the target points, and Track-Anything Model (TAM)~\cite{jmz+23} for the masks.
Drag-A-Video~\cite{yey+23} tracks the point and mask trajectories using diffusion hidden feature correspondence~\cite{tjw+23} instead of off-the-shelf models. It uses an averaged handle points feature across frames $\overline{\mathbf{h}^{\cdot}_{[\mathbf{q}^{\cdot}_i]}}=\frac{1}{F}\sum_{f=1}^F\mathbf{h}^{f}_{[\mathbf{q}^f_i]}$ as supervision to enhance temporal consistency, with the motion loss formulated as:
\begin{equation}
    \mathcal{L}_{\mathrm{motion}} = 
        \sum_{i,f} \sum_{\mathbf{q}_i^f \in \Omega(\mathbf{p}_i^f, r)}\|  \mathbf{h}^f_{[\mathbf{q}^f_i + \mathbf{d}^f_i]} - 
            \mathrm{sg}(\overline{\mathbf{h}^{\cdot}_{[\mathbf{q}^{\cdot}_i]}}) \|_1
\end{equation}
Moreover, Drag-A-Video optimizes the latent state across multiple timesteps using a shared latent state offset instead of focusing on a single timestep.

Unlike the above methods that use the DragDiffusion paradigm, VideoSwap~\cite{yyb+23} is inspired by the observation that semantic points can align the subject's shape and motion trajectory. 
It extracts the DIFT~\cite{tjw+23} feature $\mathbf{h}$ for each user-provided handle point $\{\mathbf{p}_i\}$ and adds the MLP-projected DIFT feature to the corresponding position of the diffusion encoder~\cite{cxl+24}.
These MLP projectors are trained on the diffusion loss in the surrounding region of the handle points $\Omega(\mathbf{p}_i, r)$ as follows:
\begin{equation}
    \mathcal{L} = \mathbb{E}_{
        t,\mathbf{z}_t, \epsilon_t, \mathbf{q}_i \in \Omega(\mathbf{p}_i, r)
    }{\| 
        (\epsilon_t - \epsilon_\theta (\mathbf{z}_t, t, \phi(\mathbf{h})))_{[\mathbf{q}_i]}
    \|_2^2}
\end{equation}
where $\phi(\cdot)$ indicates the trainable projection MLP.
During inference, dropping the added encoder embeddings removes the subjects at a handle point.
When moving handle points, VideoSwap utilizes the canonical atlas~\cite{KastenOWD21}, as mentioned in \cref{subsec:canonical}, to track the trajectory of handle points and target points.

\subsubsection{Pose Conditioning for Human Video Editing}\label{subsubsec:pose}
Human video editing is a domain-specific task with substantial creative and commercial potential. 
It entails translating a source human video into another temporally coherent video, featuring the target character while replicating the source video actions.
This task presents three challenges:
\begin{itemize}
    \item[$\bullet$] \textbf{Temporal Consistency}: The inter-frame smoothness is a common challenge in video modeling.
    \item[$\bullet$] \textbf{Pose Alignment}: The edited video must adhere to the source poses.
    \item[$\bullet$] \textbf{Identity Alignment}: The target video should preserve the specified character’s identity.
\end{itemize}

Follow-Your-Pose~\cite{MaHCWC0C24} focuses on investigating the pose alignment and temporal consistency challenges.
It utilizes the pose map, where vertices represent joints and edges denote limbs, as a domain-specific structural condition. This condition abstracts pose from the source video, maintaining only the structural information.
Given the scarcity of high-quality annotated human action video datasets at that time, Follow-Your-Pose decouples temporal consistency and pose alignment learning. It first learns a pose-conditioned generative model using a pose-image dataset. Subsequently, it learns temporal consistency on videos without pose annotations.
Specifically, pose maps are encoded using convolutional layers and integrated into pre-trained T2I UNet blocks~\cite{rad+22} by addition.
Follow-Your-Pose also incorporates temporal and sparse-causal attention mechanisms~\cite{wgw+23} to maintain temporal consistency.
Only the newly added parameters are trained while the pre-trained parameters remain frozen. Once training is completed, these modules are combined to enable human video editing.

DreamPose~\cite{KarrasHWK23} utilizes reference images to specify the character's identity.
Its reference image provides finer details than text prompts, such as facial and body attributes. Reference image embeddings from the image CLIP~\cite{RadfordKHRGASAM21} and VAE~\cite{dm14} encoders are merged to capture intricate details and conditioned in the diffusion model using cross-attention.
The pose conditions are concatenated with noisy latent states, expanding the input channels, as elaborated in \cref{subsec:image-diffusion}.
Moreover, it employs the hierarchical classifier-free guidance, as outlined in \cref{eq:multi-cfg}.

DisCo~\cite{wll+23} pre-trains its spatial layers on human image datasets~\cite{FuLJLQLWL22,GeZWTL19,JafarianP21,LinMBHPRDZ14} to enhance identity precision and integrates pose and background information through ControlNet~\cite{zra23}.
MotionEditor~\cite{sqz+23} addresses ghosting and blurring effects caused by conflicting poses between the source video and the reference identity. 
Instead of directly adding ControlNet~\cite{zra23} outputs to the UNet, it uses a motion adapter before integration. This adapter includes two parallel paths: one employing temporal convolutions to enhance temporal pose sequence features, and the other incorporating cross-attention and temporal self-attention blocks. The cross-attention uses pose features to query UNet features, capturing correlated motion clues in noisy latent states.

MagicPose~\cite{dyq+24} enhances identity alignment by introducing an appearance branch. This branch is duplicated from the main branch and employed to extract identity features from the reference image.
The self-attention key and value features of the main branch are augmented by concatenating with the appearance branch features: 
\begin{equation}
    \mathbf{T} = [\mathbf{W}_T \cdot \mathbf{h}, \mathbf{W}^r_T \cdot \mathbf{h}^r] \in \mathbb{R}^{d\times 2N}
\end{equation}
where $\mathbf{T} \in \{\mathbf{K}, \mathbf{V}\}$, $\mathbf{h}\in \mathbb{R}^{d\times N}$ represents the hidden features, and $\mathbf{W}_T \in \mathbb{R}^{d\times d}$ represents the projection matrices from the main branch. 
$\mathbf{h}^r\in \mathbb{R}^{d\times N}$ and $\mathbf{W}_T^r \in \mathbb{R}^{d\times d}$ are their correspondence from the appearance branch.
Similarly motivated, MagicAnimate~\cite{zjj+23} and AnimateAnyone~\cite{lxp+23} also incorporate the appearance branch. However, they distinguish themselves from MagicPose~\cite{dyq+24} by incorporating the hidden feature from the appearance branch and utilizing shared projection matrices: $\mathbf{T} = \mathbf{W}_T \cdot [\mathbf{h}, \mathbf{h}^r]\in \mathbb{R}^{d\times 2N}$.

While these advancements significantly enhance the creative potential of human video editing, they also highlight the urgent need for ethical safeguards against the misuse of identity synthesis, a topic we further discuss in \cref{sec:trend}.

\section{Benchmarking}\label{sec:bench}

\begin{figure}
    \includegraphics[width=0.99\linewidth]{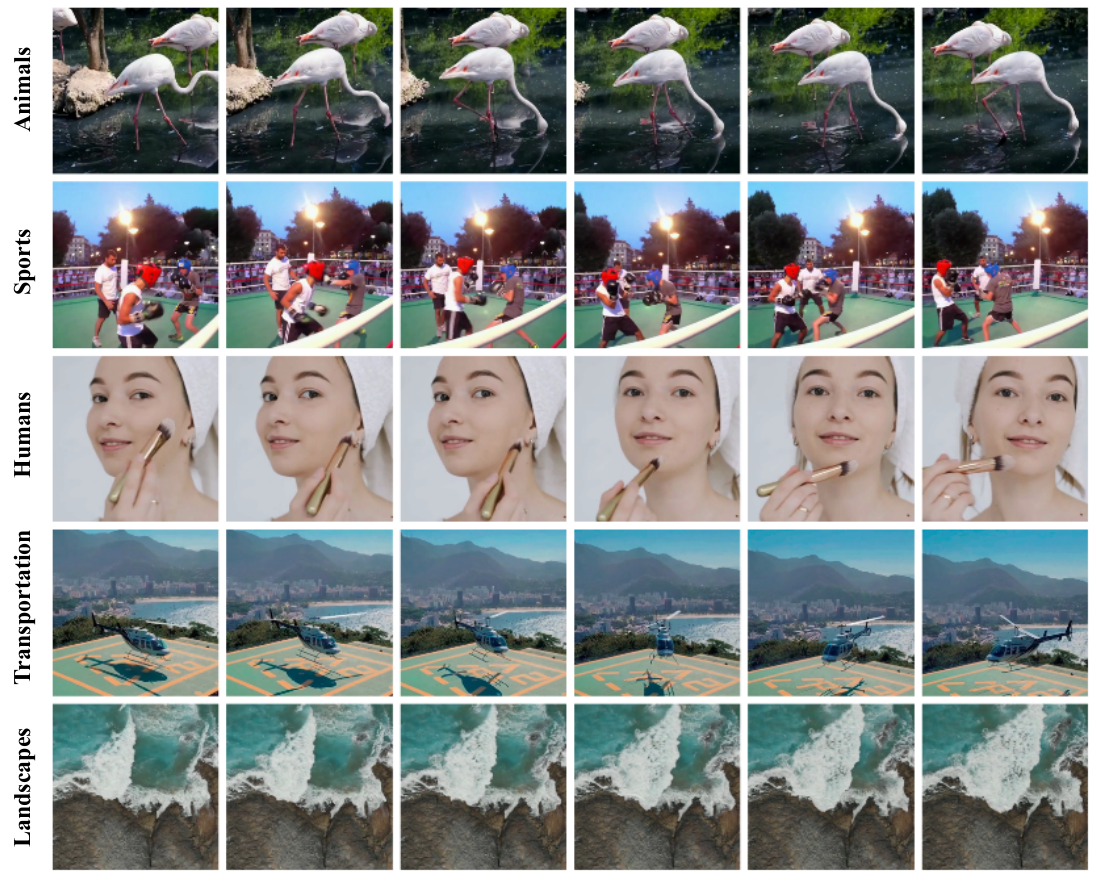}
    \caption{Some source video examples of V2VBench.}
    \label{fig:data-example}
\end{figure}

With an increasing number of research studies, there is a growing need for a benchmark to evaluate and compare their strengths and limitations, aiding in selecting the most suitable method.
Previous benchmarks either focus solely on text-to-video (T2V) \cite{ipt+23,zyj+23,yxx+23,kkx+24,mhx+24,txz+24,jgh+24} or image-to-video (I2V) \cite{zyj+23} generation without video input or lack comprehensive evaluation across various dimensions~\cite{rwy+23,jxd+23}.
This section introduces V2VBench, a benchmark that quantitatively assesses video editing methods.

We start by introducing our curated video dataset in \cref{subsec:data}, followed by the evaluation metrics in \cref{subsec:metrics}. 
Then we provide the experimental implementations in \cref{subsec:implement}.
Finally, \cref{subsec:compare} presents the quantitative evaluation results and visualizations of the 16 methods.

\begin{table*}[t]
\renewcommand{\arraystretch}{1.3}
\caption{
    V2VBench evaluation results.
    This table compares 16 diffusion-based text-guided video editing methods across eight performance metrics; higher scores indicate superior performance.
    Symbols denote specific requirements:
    $\dagger$ for test-time tuning on the source video. $\ddagger$ for video large-scaled pretraining. 
    Methods are categorized by their main technological dependencies. The best results within each category are \underline{underlined}, and the globally best metrics are highlighted in \underline{\textbf{bold font}}.
    Values in parentheses represent the standard deviation across five independent runs with different random seeds.
}
\label{tab:bench}
\centering
\resizebox{2\columnwidth}{!}{%
    \begin{tabular}{clcccccccc}
    \hline
    \begin{tabular}{@{}c@{}}\phantom{abc}\\\phantom{abcde}\end{tabular}
    & Method 
    & \makecell{Frames\\Quality} & \makecell{Video\\Quality}
    & \makecell{Semantic\\Consistency} & \makecell{Object\\Consistency}
    & \makecell{Frames Text\\Alignment} & \makecell{Video Text\\Alignment}
    & \makecell{Motion\\Alignment} & \makecell{Frames\\Pick Score} \\
    \hline
    \parbox[t]{10mm}{\multirow{5}{*}{\rotatebox[origin=c]{90}{\makecell{Network and\\Training}}}} 
    & Tune-A-Video~\cite{wgw+23} $^\dagger$ & 5.016 {\scriptsize($\pm$0.110)} & 0.449 {\scriptsize($\pm$0.071)} & 0.934 {\scriptsize($\pm$0.009)} & 0.916 {\scriptsize($\pm$0.013)} & 27.774 {\scriptsize($\pm$0.824)} & 0.257 {\scriptsize($\pm$0.012)} & -5.463 {\scriptsize($\pm$0.939)} & 20.755 {\scriptsize($\pm$0.157)} \\
    & SimDA~\cite{zqh+23} $^\dagger$ & 4.987 {\scriptsize($\pm$0.104)} & 0.480 {\scriptsize($\pm$0.080)} & 0.940 {\scriptsize($\pm$0.007)} & 0.930 {\scriptsize($\pm$0.010)} & 26.792 {\scriptsize($\pm$0.818)} & 0.249 {\scriptsize($\pm$0.013)} & -4.661 {\scriptsize($\pm$0.588)} & 20.541 {\scriptsize($\pm$0.175)} \\
    & VidToMe~\cite{xcx+23}  & \underline{5.294} {\scriptsize($\pm$0.179)} & \underline{0.597} {\scriptsize($\pm$0.045)} & \underline{0.956} {\scriptsize($\pm$0.006)} & 0.950 {\scriptsize($\pm$0.007)} & 27.550 {\scriptsize($\pm$0.922)} & 0.252 {\scriptsize($\pm$0.012)} & \underline{-2.463} {\scriptsize($\pm$0.529)} & 20.811 {\scriptsize($\pm$0.202)} \\
    & VideoComposer~\cite{WangYZCWZSZZ23} $^\ddagger$ & 4.539 {\scriptsize($\pm$0.191)} & 0.379 {\scriptsize($\pm$0.068)} & 0.919 {\scriptsize($\pm$0.013)} & 0.913 {\scriptsize($\pm$0.016)} & \underline{28.399} {\scriptsize($\pm$1.919)} & \underline{0.265} {\scriptsize($\pm$0.032)} & -7.765 {\scriptsize($\pm$0.589)} & 20.267 {\scriptsize($\pm$0.382)} \\
    & MotionDirector~\cite{ryj+23} $^\dagger$$^\ddagger$ & 5.128 {\scriptsize($\pm$0.159)} & 0.556 {\scriptsize($\pm$0.063)} & 0.952 {\scriptsize($\pm$0.010)} & \underline{\textbf{0.954}} {\scriptsize($\pm$0.009)} & 26.365 {\scriptsize($\pm$1.398)} & 0.264 {\scriptsize($\pm$0.020)} & -3.485 {\scriptsize($\pm$0.819)} & \underline{20.846} {\scriptsize($\pm$0.294)} \\
    \hline
    \parbox[t]{10mm}{\multirow{6}{*}{\rotatebox[origin=c]{90}{\makecell{Attention\\Injection}}}} 
    & Video-P2P~\cite{lzl+23} $^\dagger$ & 4.900 {\scriptsize($\pm$0.133)} & 0.382 {\scriptsize($\pm$0.069)} & 0.944 {\scriptsize($\pm$0.007)} & 0.927 {\scriptsize($\pm$0.010)} & 23.661 {\scriptsize($\pm$0.827)} & 0.190 {\scriptsize($\pm$0.011)} & -5.769 {\scriptsize($\pm$0.684)} & 19.735 {\scriptsize($\pm$0.148)} \\
    & Vid2Vid-Zero~\cite{wxl+23} & 5.089 {\scriptsize($\pm$0.048)} & 0.473 {\scriptsize($\pm$0.069)} & 0.919 {\scriptsize($\pm$0.006)} & 0.912 {\scriptsize($\pm$0.007)} & \underline{\textbf{29.012}} {\scriptsize($\pm$0.454)} & \textbf{\underline{0.274}} {\scriptsize($\pm$0.006)} & -4.047 {\scriptsize($\pm$0.361)} & \underline{20.998} {\scriptsize($\pm$0.101)} \\
    & Fate-Zero~\cite{qcz+23} & 5.033 {\scriptsize($\pm$0.021)} & 0.541 {\scriptsize($\pm$0.066)} & \underline{0.951} {\scriptsize($\pm$0.003)} & \underline{0.952} {\scriptsize($\pm$0.002)} & 25.073 {\scriptsize($\pm$0.154)} & 0.225 {\scriptsize($\pm$0.002)} & \underline{\textbf{-1.425}} {\scriptsize($\pm$0.062)} & 20.704 {\scriptsize($\pm$0.023)} \\
    & TokenFlow~\cite{mos+23} & 5.048 {\scriptsize($\pm$0.076)} & 0.552 {\scriptsize($\pm$0.083)} & 0.950 {\scriptsize($\pm$0.005)} & 0.945 {\scriptsize($\pm$0.006)} & 27.012 {\scriptsize($\pm$0.712)} & 0.247 {\scriptsize($\pm$0.010)} & -1.578 {\scriptsize($\pm$0.111)} & 20.727 {\scriptsize($\pm$0.118)} \\
    & FLATTEN~\cite{ymc+23} & \underline{5.148} {\scriptsize($\pm$0.109)} & \underline{0.571} {\scriptsize($\pm$0.047)} & 0.948 {\scriptsize($\pm$0.006)} & 0.947 {\scriptsize($\pm$0.003)} & 27.064 {\scriptsize($\pm$0.500)} & 0.250 {\scriptsize($\pm$0.006)} & -1.470 {\scriptsize($\pm$0.135)} & 20.847 {\scriptsize($\pm$0.099)} \\
    & FRESCO~\cite{syz+24} $^\dagger$ & 5.110 {\scriptsize($\pm$0.135)} & 0.558 {\scriptsize($\pm$0.080)} & 0.936 {\scriptsize($\pm$0.019)} & 0.927 {\scriptsize($\pm$0.024)} & 25.735 {\scriptsize($\pm$1.244)} & 0.231 {\scriptsize($\pm$0.017)} & -5.004 {\scriptsize($\pm$0.531)} & 20.578 {\scriptsize($\pm$0.268)} \\
    \hline
    \parbox[t]{10mm}{\multirow{5}{*}{\rotatebox[origin=c]{90}{\makecell{Latents\\Manipulation}}}} 
    & Text2Video-Zero~\cite{KhachatryanMTHW23}  & 5.131 {\scriptsize($\pm$0.076)} & 0.506 {\scriptsize($\pm$0.060)} & 0.907 {\scriptsize($\pm$0.008)} & 0.895 {\scriptsize($\pm$0.009)} & \underline{28.996} {\scriptsize($\pm$0.522)} & 0.263 {\scriptsize($\pm$0.007)} & -16.854 {\scriptsize($\pm$2.016)} & 20.573 {\scriptsize($\pm$0.117)} \\
    & Pix2Video~\cite{CeylanHM23} & 5.068 {\scriptsize($\pm$0.031)} & 0.523 {\scriptsize($\pm$0.059)} & 0.944 {\scriptsize($\pm$0.002)} & 0.942 {\scriptsize($\pm$0.003)} & 28.626 {\scriptsize($\pm$0.241)} & \underline{0.271} {\scriptsize($\pm$0.004)} & -2.926 {\scriptsize($\pm$0.128)} & \underline{\textbf{21.030}} {\scriptsize($\pm$0.046)} \\
    & ControlVideo~\cite{yyd+23}  & \underline{\textbf{5.413}} {\scriptsize($\pm$0.150)} & 0.569 {\scriptsize($\pm$0.085)} & \underline{\textbf{0.960}} {\scriptsize($\pm$0.008)} & \underline{0.946} {\scriptsize($\pm$0.011)} & 28.539 {\scriptsize($\pm$1.280)} & 0.258 {\scriptsize($\pm$0.017)} & -9.400 {\scriptsize($\pm$0.134)} & 20.931 {\scriptsize($\pm$0.297)} \\
    & Rerender~\cite{YangZLL23} & 5.029 {\scriptsize($\pm$0.246)} & \underline{\textbf{0.618}} {\scriptsize($\pm$0.121)} & 0.881 {\scriptsize($\pm$0.033)} & 0.853 {\scriptsize($\pm$0.058)} & 27.581 {\scriptsize($\pm$1.593)} & 0.259 {\scriptsize($\pm$0.018)} & -5.541 {\scriptsize($\pm$0.637)} & 20.503 {\scriptsize($\pm$0.493)} \\
    & RAVE~\cite{obh+23} & 5.060 {\scriptsize($\pm$0.036)} & 0.536 {\scriptsize($\pm$0.065)} & 0.927 {\scriptsize($\pm$0.003)} & 0.937 {\scriptsize($\pm$0.004)} & 28.441 {\scriptsize($\pm$0.382)} & 0.260 {\scriptsize($\pm$0.005)} & \underline{-2.325} {\scriptsize($\pm$0.113)} & 20.927 {\scriptsize($\pm$0.061)} \\
    \hline
    \end{tabular}
}
\end{table*}

\subsection{Video Editing Dataset}\label{subsec:data}
Most video editing methods use a triplet of inputs: a source video, a descriptive text prompt outlining the source video, and a target text prompt specifying the desired edits.
In V2VBench, we curated 50 video clips and annotated each with 1 descriptive and 3 target text prompts, organizing 150 editing tasks.

\begin{figure}
\centering
\subfloat[Videos categories.]{
    \includegraphics[width=.45\linewidth]{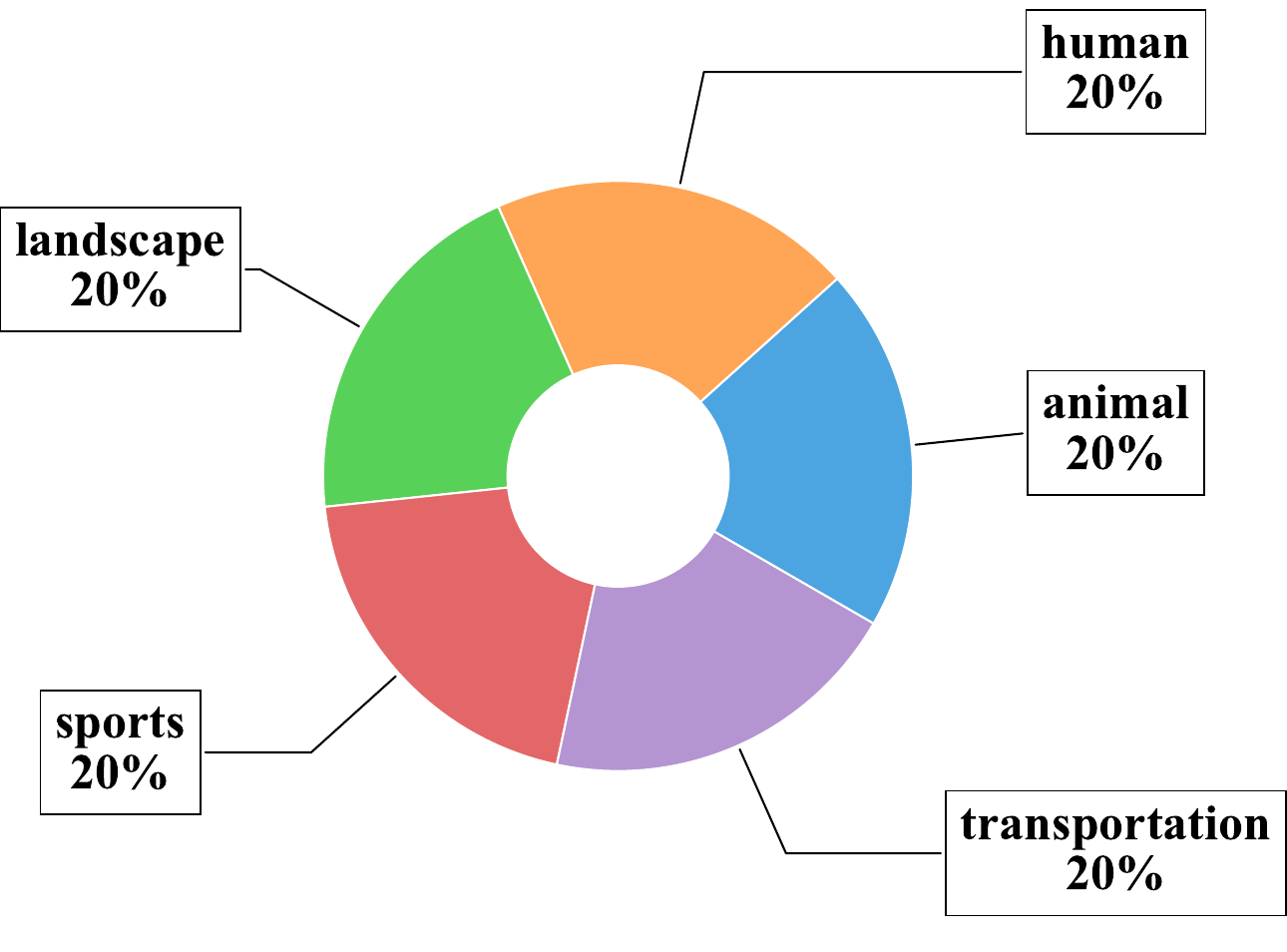}
    \label{subfig:bench-category}
}
\hfil
\subfloat[Videos dynamic range.]{
    \includegraphics[width=.45\linewidth]{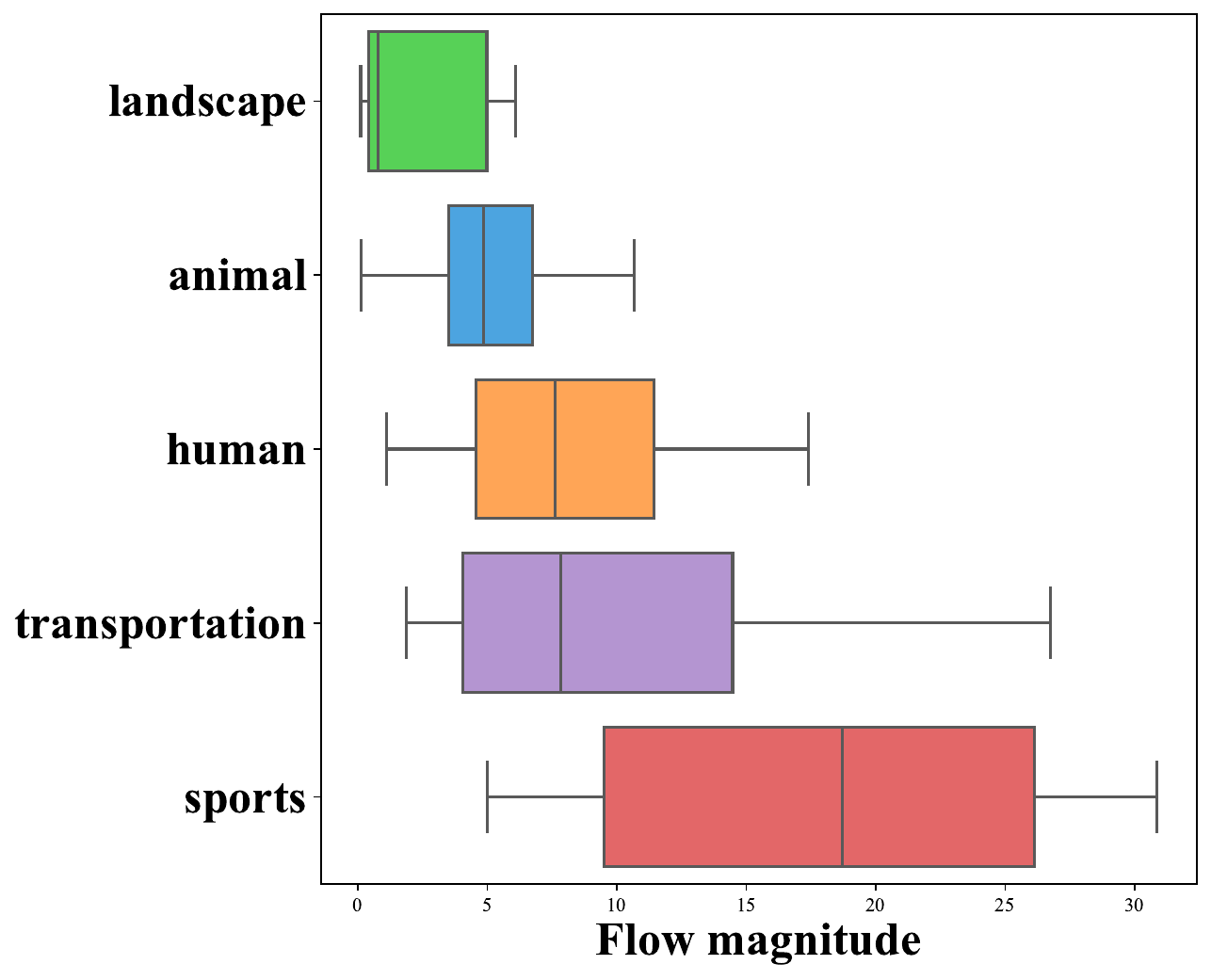}
    \label{subfig:bench-motion}
}
\hfil
\subfloat[Target prompts categories.]{
    \includegraphics[width=.6\linewidth]{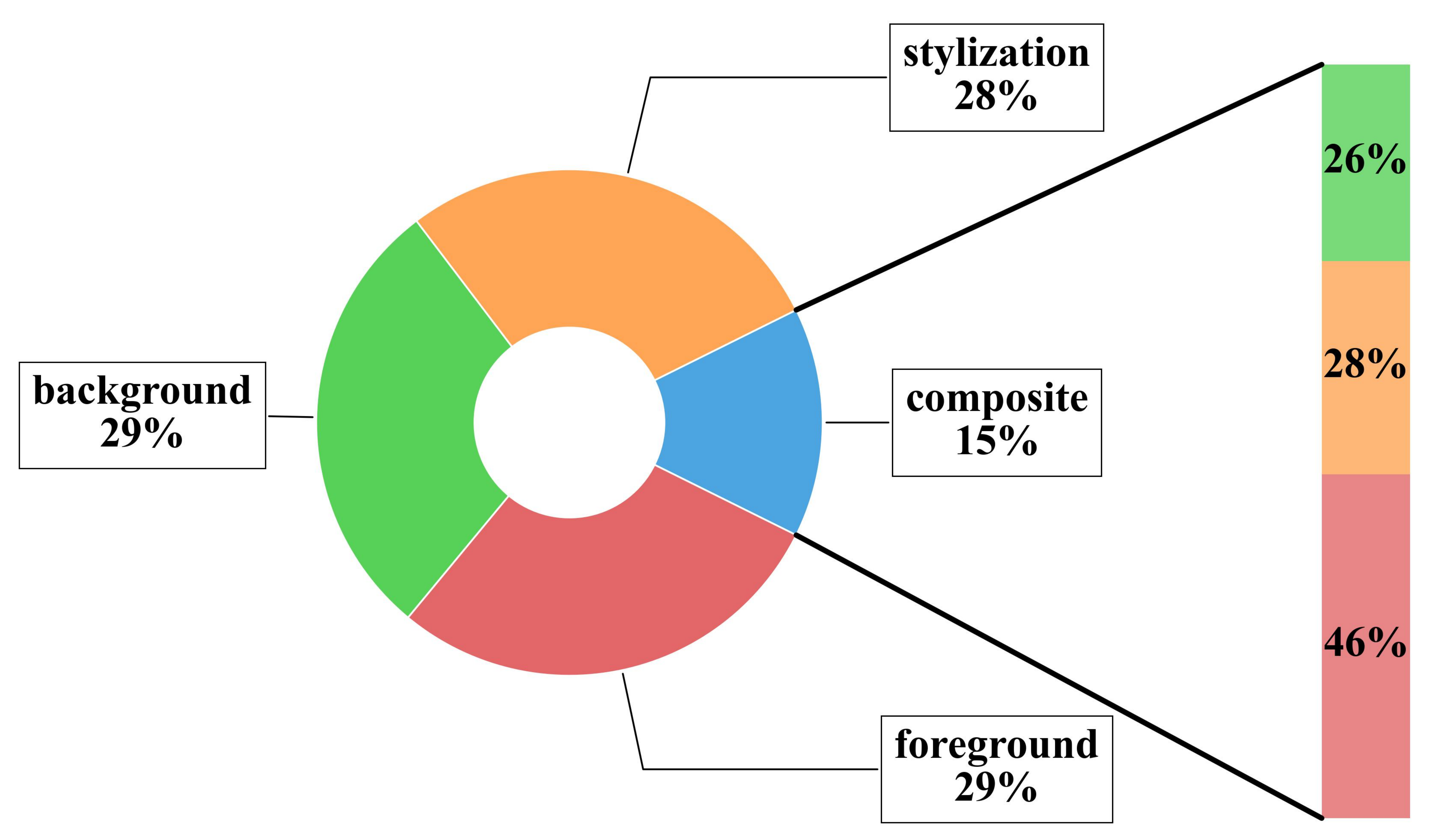}
    \label{subfig:bench-prompt}
}
\hfil
\subfloat[Target words cloud.]{
    \includegraphics[width=.3\linewidth]{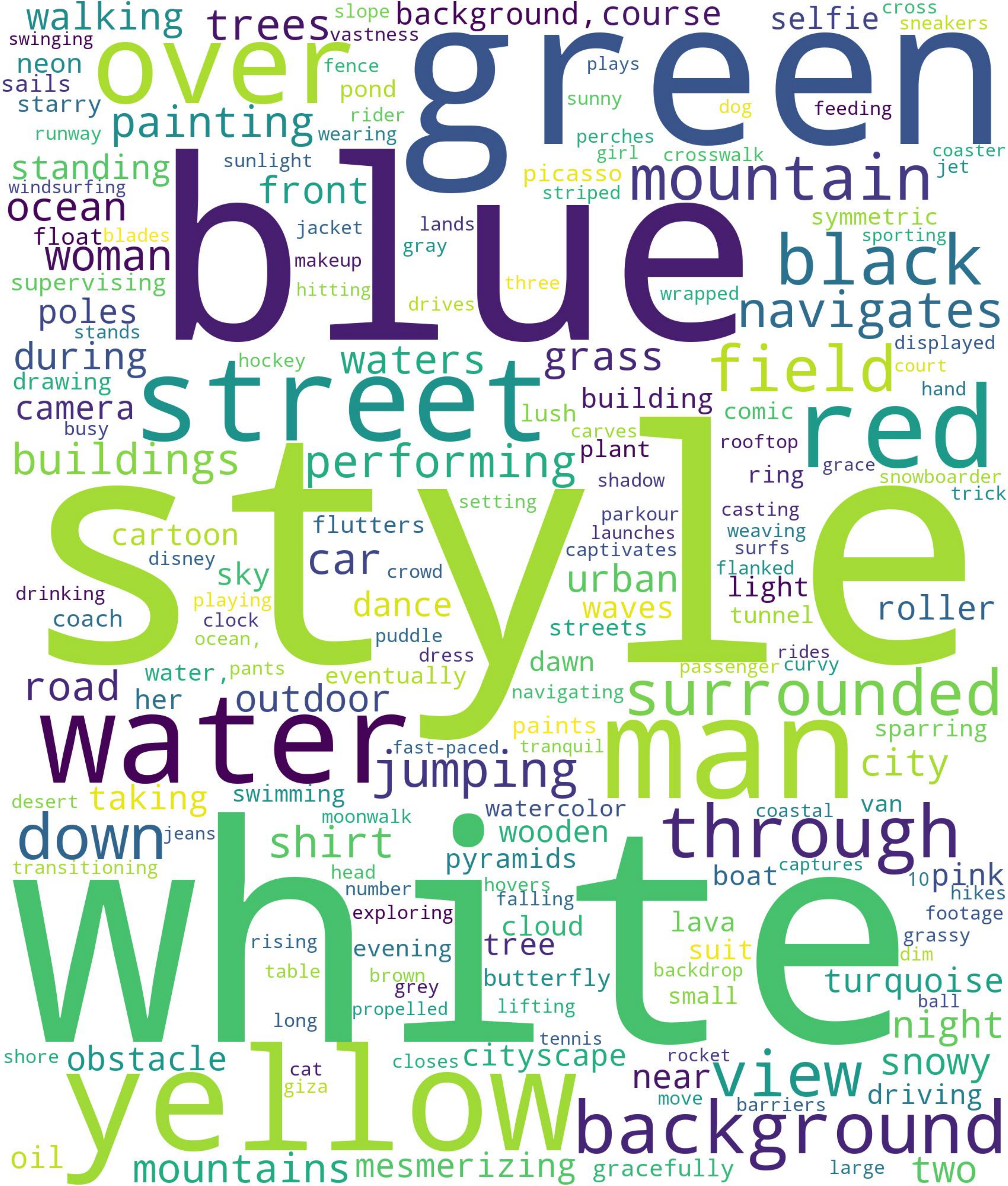}
    \label{subfig:bench-wordcloud}
}

\caption{Statistics of V2VBench videos and prompts.}
\label{fig:bench-data}
\end{figure}

\vspace{6pt} \noindent$\bullet$~\textbf{Source videos.}
We collect high-quality, open-licensed clips from the internet as the source videos for V2VBench. These clips range from 2 to 100 seconds in length, with an average of 30 frames per second (FPS) and various aspect ratios, all with a minimum resolution of $512\times512$ pixels. The videos are evenly categorized into five common categories relevant to real application scenarios: animals, sports, humans (non-sports), landscapes, and transportation, as shown in \cref{fig:data-example,subfig:bench-category}.

To maintain compatibility with most methods' configurations, we sample 24 frames from each video clip. 
The sampling stride is manually adjusted to maintain a balanced dynamic range, ensuring adequate coverage.
\cref{subfig:bench-motion} validates this by measuring the dense optical flow magnitude. Landscape videos show the lowest average dynamic range, while sports videos demonstrate the highest, consistent with expectations.
Additionally, we downscale and crop these video frames to a $512\times512$ spatial resolution, ensuring the subjects are relatively centered in most frames.

\vspace{6pt} \noindent$\bullet$~\textbf{Descriptive Text Prompts.}
We utilize the video captioning approach outlined in SVD~\cite{ats+23} to generate descriptive prompts for video clips. First, we employ two caption synthesis models: the image captioning model CoCa~\cite{YuWVYSW22} annotates the first frame of each video clip, while VideoBLIP~\cite{kzf+23} provides the video-level caption, effectively capturing spatial and temporal aspects of the video, respectively. These descriptions are integrated using the language model \cite{gpt4o} to leverage both visual aspects.

\vspace{6pt} \noindent$\bullet$~\textbf{Target Text Prompts.}
To cover the diverse application scenarios of video editing, we create three target text prompts for each video clip featuring different editing types, including foreground object editing, background editing, and stylization. Inspired by BlanceCC~\cite{rwy+23}, we also incorporate composite edits that combine multiple simple edits, presenting more challenging tasks.

We manually correct discrepancies and simplify overly complicated entries. Consequently, simple edits are nearly evenly distributed, covering approximately 29\% of the target prompts, while composite edits account for 15\%, as depicted in~\cref{subfig:bench-prompt}.
The word cloud generated from the target prompt is shown in \cref{subfig:bench-wordcloud}.

\subsection{Evaluation Metrics}\label{subsec:metrics}
The lack of universally accepted automatic metrics leads existing methods to use diverse evaluation indicators, complicating a comprehensive understanding of their advantages and limitations. Therefore, we employ eight evaluation metrics to provide a well-rounded assessment.

\vspace{6pt} \noindent$\bullet$~\textbf{Frames Quality.}
Before considering all frames collectively, the quality of each frame constitutes the fundamental basis for determining the overall video quality.
We utilize the LAION aesthetic predictor~\cite{crr+21}, calibrated based on human rankings, to assess the edited video frames. This predictor evaluates subjective aspects including layout, richness, artistry, and visual appeal~\cite{zyj+23}. Subsequently, we compute the average aesthetic scores to derive the overall quality score of the edited video.

\vspace{6pt} \noindent$\bullet$~\textbf{Video Quality.}
Alongside frame-by-frame evaluation, we employ the DOVER score~\cite{Wu0LCHWSYL23} for video-level assessment.
DOVER is the state-of-the-art video quality assessment method trained on a large-scale human-ranked video dataset. In addition to aesthetic considerations, the DOVER score evaluates technical aspects including artifacts, distortion, blurring, and meaningfulness.

\vspace{6pt} \noindent$\bullet$~\textbf{Semantic Consistency.}
CLIP~\cite{RadfordKHRGASAM21} vision embeddings are widely used to capture image semantic information.
The cosine similarity between adjacent frames' CLIP embeddings is a standard metric for assessing inter-frame consistency and the overall smoothness of the edited video.

\vspace{6pt} \noindent$\bullet$~\textbf{Object Consistency.}
Beyond evaluating semantic consistency, we examine whether object-level appearance remains consistent in the edited videos. We employ DINO~\cite{CaronTMJMBJ21,mtt+23}, a self-supervisely pre-trained image embedding model, to compute object-level inter-frame similarity.

\vspace{6pt} \noindent$\bullet$~\textbf{Frames Text Alignment.}
In addition to assessing the visual quality of the edited videos, alignment with the target text prompts is crucial for controllable video manipulation. CLIP score~\cite{RadfordKHRGASAM21} is the most widely used metric for evaluating vision-text alignment. Following established methods, we compute the CLIP score frame by frame and aggregate the scores by averaging across all frames.

\vspace{6pt} \noindent$\bullet$~\textbf{Video Text Alignment.}
Isolated frames cannot fully represent motion, and the original CLIP score may not accurately capture video-level alignment. We also compute the ViCLIP score~\cite{yyy+23}. Trained on video-text pairs, the ViCLIP score assesses alignment between spatio-temporal content and target prompts.

\vspace{6pt} \noindent$\bullet$~\textbf{Motion Alignment.}
Maintaining consistent motion alignment with the input source video is crucial. To evaluate this, we utilize optical flow computed by GMFlow~\cite{XuZ0RT22} for each source-editing video pair and measure their difference, also known as endpoint error (EPE).
To align with other metrics, which aim for higher values indicating better performance, we negate the final EPE.

\vspace{6pt} \noindent$\bullet$~\textbf{Frames Pick Score.}
Human preferences are sometimes complex and cannot be fully captured by feature-level similarity alone. The Pick Score~\cite{KirstainPSMPL23} is a CLIP-based scoring function trained on large-scale human preference annotations, considering both image and text inputs to reflect content quality and alignments. We calculate the frame-wise Pick Score and average them as a metric.

\vspace{6pt} \noindent$\bullet$~\textbf{Memory Usage and Running Time.}
Besides quality considerations, efficiency metrics such as peak memory usage and running time underscore the strengths and weaknesses of methods.
An ideal method should run efficiently on consumer-level devices. Peak memory usage could pose a barrier in broader usage scenarios, making it a crucial efficiency metric to monitor.
Running time is categorized into preprocessing time and generation time.
Preprocessing for a source video is performed only once and does not need to be repeated for each subsequent editing of the same video. It includes depth map extraction, latent inversion, and sample-specific fine-tuning.
Generation time is required for each editing instance, such as diffusion model sampling.

\begin{figure}
\includegraphics[width=0.99\linewidth]{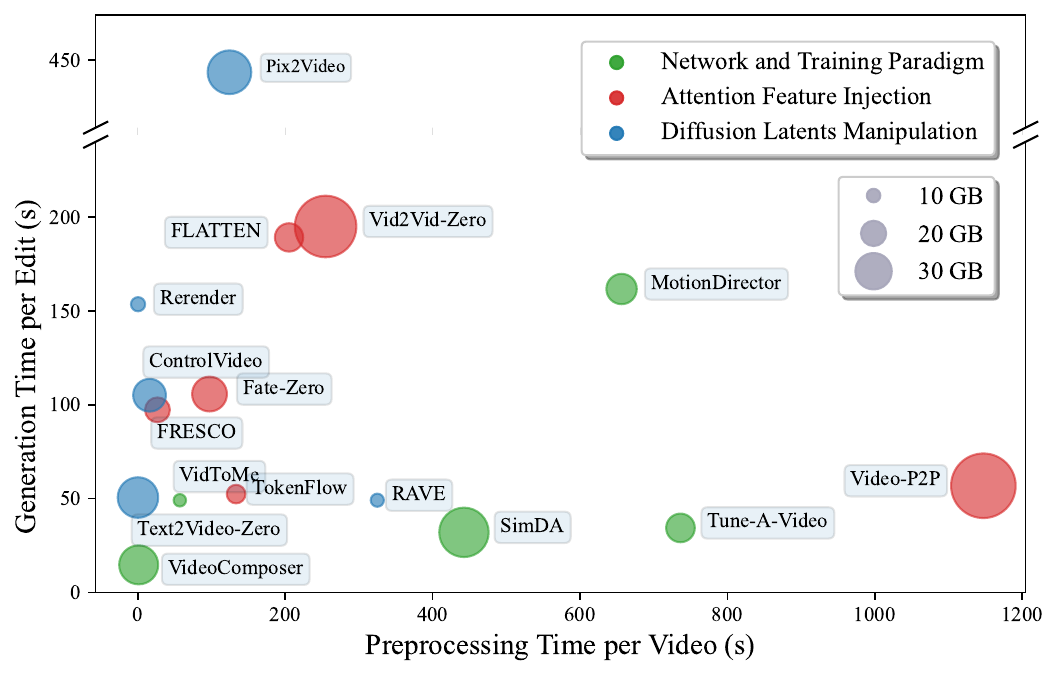}
\caption{
    Memory usage and running time comparison of 16 diffusion model-based video editing methods. The color represents the category, and the radius indicates peak memory usage.
}
\label{fig:cost}
\end{figure}

\begin{figure*}
\includegraphics[width=0.99\linewidth]{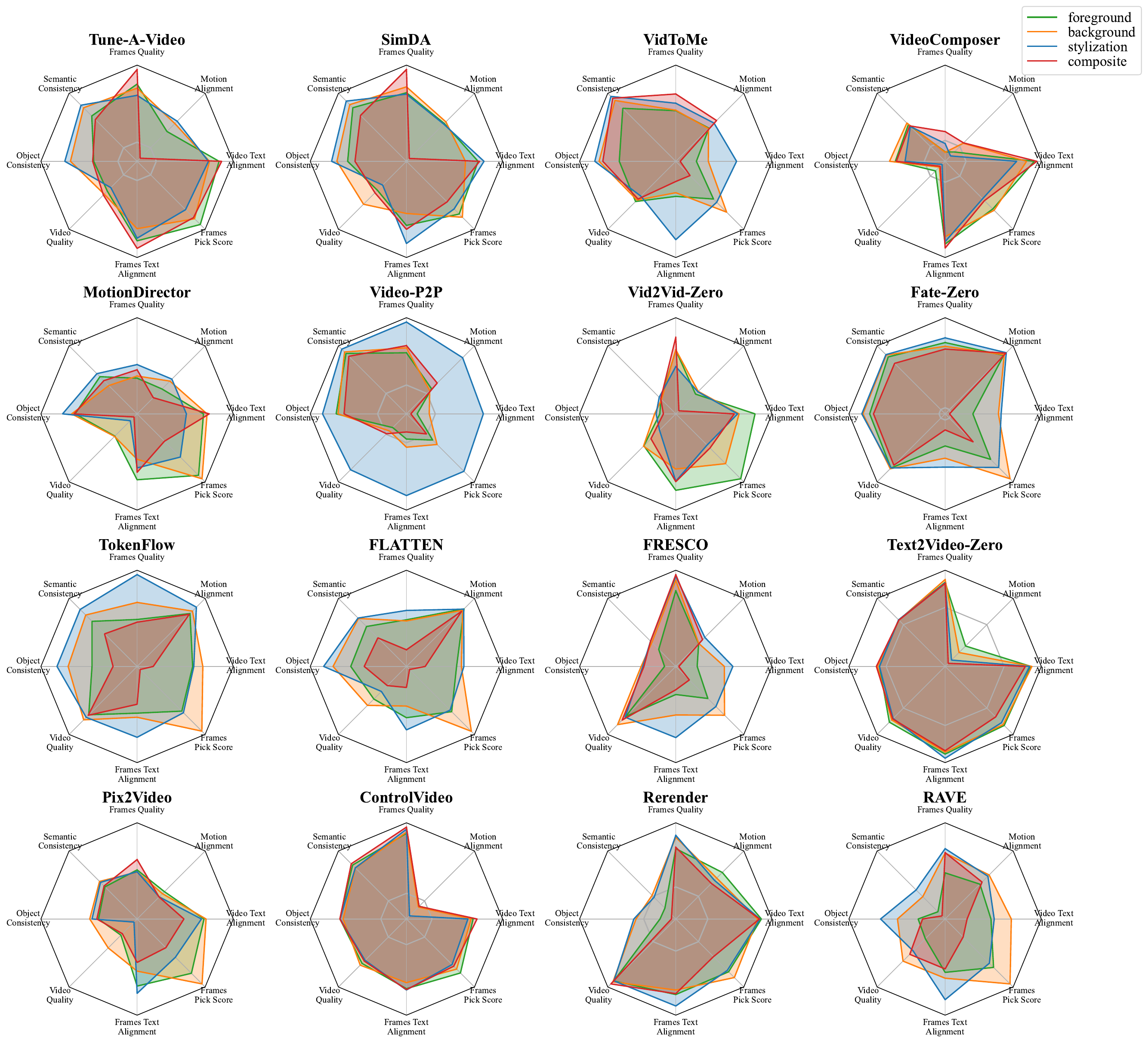}
\caption{
    Further comparisons on V2VBench. 
    Each color represents the metrics for different editing types. 
    Please refer to the appendix for additional details and normalization methods.
}
\label{fig:bytask}
\end{figure*}

\begin{table*}[t]
\renewcommand{\arraystretch}{1.3}
\caption{
    Quantitative evaluation on V2VBench using Stable Diffusion (SD) \cite{rad+22} v1-5 and v2-1 backbones.
    While adopting SD v2-1 generally yields marginal performance gains across most scenarios, these improvements are largely statistically insignificant. Overall, method performance remains robust and consistent across both backbone architectures.
}
\label{tab:pretrained}
\centering
\resizebox{2\columnwidth}{!}{%
    \begin{tabular}{llcccccccc}
    \hline
    Method 
    & Pretrained
    & \makecell{Frames\\Quality} & \makecell{Video\\Quality}
    & \makecell{Semantic\\Consistency} & \makecell{Object\\Consistency} 
    & \makecell{Frames Text\\Alignment} & \makecell{Video Text\\Alignment} 
    & \makecell{Motion\\Alignment} & \makecell{Frames\\Pick Score}\\
    \hline
    \multirow{2}{*}{Tune-A-Video~\cite{wgw+23}}
    & SD v1-5 & 5.016 {\scriptsize($\pm$0.110)} & 0.449 {\scriptsize($\pm$0.071)} & 0.934 {\scriptsize($\pm$0.009)} & 0.916 {\scriptsize($\pm$0.013)} & 27.774 {\scriptsize($\pm$0.824)} & 0.257 {\scriptsize($\pm$0.012)} & -5.463 {\scriptsize($\pm$0.939)} & 20.755 {\scriptsize($\pm$0.157)} \\
    & SD v2-1 & 5.020 {\scriptsize($\pm$0.098)} & 0.486 {\scriptsize($\pm$0.051)} & 0.935 {\scriptsize($\pm$0.009)} & 0.920 {\scriptsize($\pm$0.011)} & 28.186 {\scriptsize($\pm$0.705)} & 0.261 {\scriptsize($\pm$0.010)} & -5.429 {\scriptsize($\pm$0.803)} & 20.739 {\scriptsize($\pm$0.161)} \\
    \hline
    \multirow{2}{*}{Vid2Vid-Zero~\cite{wxl+23}}
    & SD v1-5 & 5.089 {\scriptsize($\pm$0.048)} & 0.473 {\scriptsize($\pm$0.069)} & 0.919 {\scriptsize($\pm$0.006)} & 0.912 {\scriptsize($\pm$0.007)} & 29.012 {\scriptsize($\pm$0.454)} & 0.274 {\scriptsize($\pm$0.006)} & -4.047 {\scriptsize($\pm$0.361)} & 20.998 {\scriptsize($\pm$0.101)} \\
    & SD v2-1 & 5.107 {\scriptsize($\pm$0.012)} & 0.510 {\scriptsize($\pm$0.008)} & 0.920 {\scriptsize($\pm$0.004)} & 0.910 {\scriptsize($\pm$0.002)} & 29.008 {\scriptsize($\pm$0.084)} & 0.277 {\scriptsize($\pm$0.001)} & -4.368 {\scriptsize($\pm$0.137)} & 21.006 {\scriptsize($\pm$0.011)} \\
    \hline
    \multirow{2}{*}{Fate-Zero~\cite{qcz+23}}
    & SD v1-5 & 5.033 {\scriptsize($\pm$0.021)} & 0.541 {\scriptsize($\pm$0.066)} & 0.951 {\scriptsize($\pm$0.003)} & 0.952 {\scriptsize($\pm$0.002)} & 25.073 {\scriptsize($\pm$0.154)} & 0.225 {\scriptsize($\pm$0.002)} & -1.425 {\scriptsize($\pm$0.062)} & 20.704 {\scriptsize($\pm$0.023)} \\
    & SD v2-1 & 5.030 {\scriptsize($\pm$0.016)} & 0.574 {\scriptsize($\pm$0.008)} & 0.951 {\scriptsize($\pm$0.002)} & 0.951 {\scriptsize($\pm$0.002)} & 25.115 {\scriptsize($\pm$0.122)} & 0.225 {\scriptsize($\pm$0.001)} & -1.411 {\scriptsize($\pm$0.060)} & 20.718 {\scriptsize($\pm$0.016)} \\
    \hline
    \multirow{2}{*}{FLATTEN~\cite{ymc+23}}
    & SD v1-5 & 5.148 {\scriptsize($\pm$0.109)} & 0.571 {\scriptsize($\pm$0.047)} & 0.948 {\scriptsize($\pm$0.006)} & 0.947 {\scriptsize($\pm$0.003)} & 27.064 {\scriptsize($\pm$0.500)} & 0.250 {\scriptsize($\pm$0.006)} & -1.470 {\scriptsize($\pm$0.135)} & 20.847 {\scriptsize($\pm$0.099)} \\
    & SD v2-1 & 5.191 {\scriptsize($\pm$0.010)} & 0.584 {\scriptsize($\pm$0.007)} & 0.949 {\scriptsize($\pm$0.002)} & 0.947 {\scriptsize($\pm$0.001)} & 27.077 {\scriptsize($\pm$0.086)} & 0.250 {\scriptsize($\pm$0.001)} & -1.512 {\scriptsize($\pm$0.041)} & 20.883 {\scriptsize($\pm$0.009)} \\
    \hline
    \end{tabular}
}
\end{table*}

\subsection{Experiment Implementation}\label{subsec:implement}
To ensure a fair and unbiased comparison, we exclude methods that heavily rely on off-the-shelf external image editing methods (\eg LAMP~\cite{rlt+23} and CCEdit~\cite{rwy+23}) or require prompts beyond text (\eg interactive point-based methods). 
Furthermore, methods with prohibitive computational overhead, such as LNA-based approaches requiring more than two hours per-video preprocessing durations, are omitted to ensure practical benchmarking.

All experiments are conducted on a single NVIDIA A100 80GB GPU. All methods use StableDiffusion~\cite{rad+22} v1-5~\footnote{https://huggingface.co/runwayml/stable-diffusion-v1-5} as the pre-trained foundation model, except for VideoComposer~\cite{WangYZCWZSZZ23} and MotionDirector~\cite{ryj+23}, which require video-pretrained models. 
We uniformly use ControlNet~\cite{zra23} v1-1-depth~\footnote{https://huggingface.co/lllyasviel/ControlNet-v1-1} and MiDaS~\cite{RanftlLHSK22} to extract the depth map for methods relying on structural information. 
To account for stochastic variance across different runs, we performed five samples per edit using distinct random seeds.
We use the negative prompt "long body, low-res, bad anatomy, bad hands, missing fingers, extra digit, fewer digits, cropped, worst quality, low quality, deformed body, bloated, ugly, blurry, low res, unaesthetic" when it is applicable for the methods.
All methods run with xFormer~\cite{xFormers2022} enabled to avoid out-of-memory issues. 
Other configurations adhere to those specified in the original repositories of the respective methods. We follow the code in cases where code configurations differ from those reported in the papers.

\subsection{Comparative Analysis}\label{subsec:compare}
We reimplement 16 methods utilizing their officially released code repositories and weights (if available) on the 50 standardized video clips mentioned in \cref{subsec:data}.
\cref{tab:bench}~presents the quality metrics and \cref{fig:cost}~presents the visualization of efficiency metrics.
We aim to provide general guidelines for method selection across application scenarios.

\vspace{6pt} \noindent
$\bullet$~\textbf{No single method exhibits dominance across all metrics}, underlining the necessity of a multi-dimensional performance analysis.
No individual framework secures the top ranking in more than three metrics, highlighting the current architectural trade-offs in video generation.

\vspace{6pt} \noindent
$\bullet$~\textbf{Feature injection methods exhibit superior performance in motion alignment.}
Regarding the motion alignment metric, even the third-ranked feature injection approach outperforms the leading methods from all other categories.
Most techniques in this category explicitly leverage optical flow to guide feature injection for cross-frame interaction. This ensures that the edited video maintains motion trajectories highly consistent with the source input.
Consequently, when the primary objective is to preserve source motion, particularly in scenarios involving large-scale or unstable movement, feature injection methods represent the most robust architectural choice.
Conversely, this emphasis on motion alignment may lead to degraded visual quality compared to other categories. This performance gap stems from potential distribution shifts induced by internal feature injection, representing a fundamental trade-off between temporal consistency and frame-wise fidelity.

\vspace{6pt} \noindent
$\bullet$~\textbf{Diffusion latent manipulation methods have shown advences on visual quality.}
Leading diffusion latent manipulation methods, such as ControlVideo~\cite{yyd+23} and Rerender~\cite{YangZLL23}, demonstrate superior visual fidelity and temporal smoothness compared to existing benchmarks.
This performance gain stems from performing motion alignment primarily within the latent space, rather than through weight fine-tuning or feature-level modification.
By mitigating the risk of distribution drift, these models leverage their inherent generative priors for high-quality frame synthesis instead of contending with underfitted motion modeling. Consequently, these approaches are preferred when maximizing visual quality is the primary objective.

\vspace{6pt} \noindent$\bullet$~\textbf{Certain methods exhibit performance disparities across different tasks.}
\cref{fig:bytask} indicates Tune-A-Video~\cite{wgw+23} and SimDA~\cite{zqh+23} demonstrate an evident decline in motion alignment for more challenging composited editing tasks. Video-P2P~\cite{lzl+23} exhibits greater proficiency in stylization compared to other tasks. Fate-Zero~\cite{qcz+23} emphasizes frame consistency but exhibits relative weakness in text alignment, particularly for more challenging composited editing tasks. The results of Rerender~\cite{YangZLL23} suggest that it complements the weaknesses of Fate-Zero. Text2Video-Zero~\cite{KhachatryanMTHW23} and ControlVideo~\cite{yyd+23} perform consistently well across all tasks and metrics, except for motion alignment with the source video.
Further research should explore ways to maintain high-level performance across different tasks.

\vspace{6pt} \noindent$\bullet$~\textbf{Efficiency is highly dependent on the methods' design.}
Although feature injection and diffusion latent manipulation methods may slightly increase generation time per edit, they typically require significantly less preprocessing time. 
In contrast, test-time tuning-based methods$^\dagger$ generally demand substantially higher preprocessing time and memory costs.
VidToMe~\cite{xcx+23}, TokenFlow~\cite{mos+23}, Rerender~\cite{YangZLL23}, and RAVE~\cite{obh+23} exhibit superiority in running efficiency and are compatible with many consumer-level devices while delivering satisfactory performance on several metrics.

\subsection{Impact of Pretrained Diffusion Backbones}
Most existing methods rely on Text-to-Image diffusion backbones. As detailed in \cref{subsec:implement}, we standardized our benchmarking using Stable Diffusion (SD) v1-5 to ensure fairness.
In this section, we investigate the impact of utilizing different pretrained backbones.
We selected methods that exhibit competitive performance across specific metrics and rely exclusively on SD-based architectures.
Consequently, approaches utilizing alternative backbones (\eg, MotionDirector~\cite{ryj+23}, which relies on ModelScopeT2V~\cite{ModelScope}) were excluded.
Ultimately, we evaluate four representative methods, Tune-A-Video~\cite{wgw+23}, Vid2Vid-Zero~\cite{wxl+23}, Fate-Zero~\cite{qcz+23}, and FLATTEN~\cite{ymc+23}, with results summarized in \cref{tab:pretrained}.

Empirically, switching between backbones does not yield significant performance fluctuations for any given method; while the SD v2-1 series exhibits a marginal superiority, the gains remain statistically insignificant.
This consistency underscores the robustness of the evaluated methods across model iterations, provided there is no fundamental architectural shift.
Furthermore, we observed that SD v2-1 consistently produces lower variance across multiple runs. This reduced stochasticity may stem from the diminished generative diversity inherent to SD v2-1, a phenomenon widely recognized within the community and often attributed to the more stringent content filtering applied to its training dataset.

\section{Challenges and Emerging Trends}\label{sec:trend}
\begin{figure*}
\includegraphics[width=0.99\linewidth]{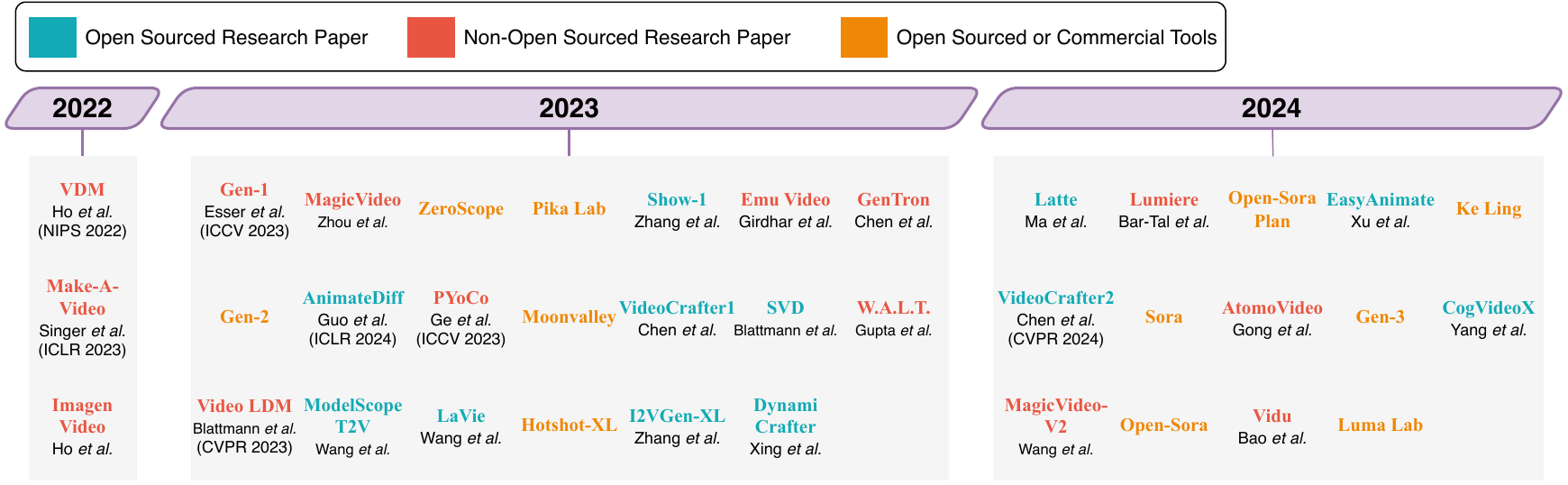}
\caption{
    Chronological overview of recent foundational video diffusion models and tools.
}
\label{fig:foundation-models}
\end{figure*}

Generative methods based on diffusion models have become widely accepted solutions for video editing, with significant advancements achieved in both academic research and commercial applications. 
Nonetheless, their full potential in video editing remains underexplored. This section outlines unresolved issues and promising opportunities for future exploration.

\vspace{6pt} \noindent$\bullet$~\textbf{Video Dataset.}
The success of image generation and editing is primarily due to the availability of large-scale, high-quality image datasets.
However, commonly used video datasets are relatively small and low quality~\cite{ats+23} because collecting, filtering, and storing video samples are significantly more challenging.
Additionally, most available video content is derived from real-world scenarios, offering less fantastical and imaginative content than images.
Although new video datasets are being developed~\cite{XueHZS00FG22,yyy+23,wy24,taw+24}, their performance in generating fantastical and imaginative content is still inferior to image counterparts.
An ideal dataset may need to integrate data from multiple sources, such as images, videos, 3D, and even 4D content.

\vspace{6pt} \noindent$\bullet$~\textbf{Foundational Video Models.}
Recent breakthroughs have highlighted the success of a paradigm where foundational models are first trained on large-scale datasets with extensive iterations, followed by post-training or fine-tuning to develop innovative methods.
However, training these foundational models remains an unaffordable burden for most researchers. This challenge is especially pronounced in video modalities, where the intrinsic complexity demands extensive data collection, cleaning, storage, and significantly higher training costs.
As shown in \cref{fig:foundation-models}, most early research on foundational video diffusion models did not make their weights publicly accessible. As a result, earlier video editing studies \cite{wgw+23,zqh+23,mos+23,obh+23} relied on foundational image diffusion models, employing training tricks or feature manipulation techniques to approximate dynamic capabilities. These methods, however, suffer from limited video training, leading to inconsistencies in generalization.
More recent video editing approaches \cite{ryj+23,WangYZCWZSZZ23} have leveraged open-source video foundational models \cite{jhd+23}, improving frame consistency and generalizability. Yet, these open-source models still lag behind the top commercial tools \cite{pika,sora,luma,gen3} and occasionally produce unavoidable artifacts. Advancing open-source foundational video diffusion models in the future could significantly foster video editing research.

\vspace{6pt} \noindent$\bullet$~\textbf{Efficiency.}
The inference time for diffusion models is a wildly recognized constraint compared to other generative models, such as GANs~\cite{mkk+18} and VAEs~\cite{dm14,ok17,prb21}, which only require a single forward pass. Accelerating the reverse diffusion process is an active area of research and remains an unsolved challenge~\cite{ada+23,SongD0S23,MengRGKEHS23,tmr+23,syl+23,ctw+24}. 
Backbone design also significantly impacts both training and inference efficiency. Despite efforts to design efficient architectures~\cite{wgw+23,zqh+23,PeeblesX23,smj+23,xyg+24}, further reducing processing time and memory usage remains a valuable topic for exploration.

\begin{figure}
\includegraphics[width=0.99\linewidth]{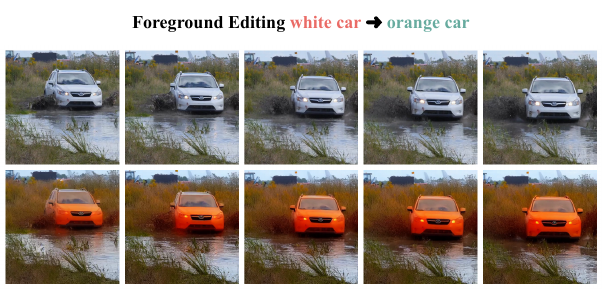}
\caption{
    \textbf{Editing spillover failure case}: The prompt intends to change the car's color from white to orange, but the surrounding reed is unexpectedly modified, resulting in artifacts.
}
\label{fig:spillover}
\end{figure}

\vspace{6pt} \noindent$\bullet$~\textbf{Editing Precision.}
A common problem with current editing methods is editing spillover, where target attributes intended for one object inadvertently affect other objects that should remain unchanged, as shown in \cref{fig:spillover}. Beyond the text and point prompts discussed in this paper, providing fine-grained spatio-temporal control remains challenging. Developing precise prompting formats that avoid complex interactions is essential for improving control accuracy. Additionally, exploring multimodal prompts offers the potential for greater flexibility.

\vspace{6pt} \noindent$\bullet$~\textbf{Evaluation.}
Currently, widely used metrics predominantly rely on image-based models rather than video-based ones.
The importance of alignment in video tasks is often overlooked.
Furthermore, employing multiple metrics for comprehensive coverage can be cumbersome and may not consistently align with human intuition when assessed individually. A unified metric is necessary to simplify comparisons and better capture intuitive assessments.
While V2VBench facilitates benchmarking for general video editing, it does not currently encompass specific interaction scenarios, such as pointing-based or human-pose conditioning in \cref{subsec:control}. These conditional modalities offer more proactive and intuitive user experiences; however, due to the scope of this survey, we defer the exploration of these specialized applications, along with expanded benchmarking scales, to future research.

\vspace{6pt} \noindent
$\bullet$~\textbf{Ethical Considerations and Deepfakes.}
The high-fidelity manipulation of human attributes, particularly the identity and pose alignment video editing techniques discussed in \cref{subsubsec:pose}, introduces societal risks, most notably the proliferation of deepfakes~\cite{deepfake}.
The democratization of these tools facilitates the creation of non-consensual content, unauthorized identity synthesis, and the spread of visual misinformation, which poses a direct threat to digital trust and security.
As editing capabilities move toward real-time and zero-shot performance, the distinction between authentic and generated media becomes increasingly blurred.
Consequently, the development of technical countermeasures, such as robust digital watermarking and detection models \cite{deepfake-1,deepfake-2,deepfake-3,deepfake-4}, must be prioritized alongside generative performance.
Future research should treat ethical safety not as a post-hoc consideration but as a fundamental architectural constraint in the design of video foundational models.

\section{Conclusion}\label{sec:conclusion}
This paper extensively dives into diffusion model-based video editing methods, providing detailed formulations and explanations of their intrinsic dependencies. It addresses the gap left by previous studies, which often treated video editing as an accessory task with simplistic descriptions.
We categorize video editing methods based on their core technologies: network architecture innovation, feature injection, diffusion, video representation, and control signals. Additionally, we outline the evolutionary trajectory of methods within each category.
Furthermore, we introduce V2VBench, which comprehensively evaluates 150 edits across 4 editing tasks using 10 metrics.
We compare 16 methods on V2VBench to provide comprehensive insights.
Finally, we address research challenges and outline promising directions for further exploration.

\begin{acknowledgements}
This research / project is supported by the National Research Foundation, Singapore, and Cyber Security Agency of Singapore under its National Cybersecurity R\&D Programme and CyberSG R\&D Cyber Research Programme Office. Any opinions, findings and conclusions or recommendations expressed in these materials are those of the author(s) and do not reflect the views of National Research Foundation, Singapore, Cyber Security Agency of Singapore as well as CyberSG R\&D Programme Office, Singapore.
\end{acknowledgements}

\section*{Conflict of interest}
The authors have no conflict of interest to declare relevant to this article's content.

\section*{Data Availability Statement}
The V2VBench dataset, introduced in \cref{sec:bench} for benchmarking purposes, is available in the repository~\url{https://github.com/wenhao728/awesome-diffusion-v2v}.

\bibliographystyle{spmpsci}      

\bibliography{egbib}   

\end{document}